%% file: preprint.tex
\documentclass{vmax-preprint}

\usepackage{nicefrac}
\usepackage{tikz}
\usetikzlibrary{arrows.meta, positioning, shapes}

\graphicspath{{figs_vmax/}{figs/}}

\input{_numbers}

\input{_numbers_math}

\title{PopuLoRA: Co-Evolving LLM Populations for Reasoning Self-Play}

\author{Roger Creus Castanyer}
\author{Geoffrey Bradway}
\author{Lorenz Wolf}
\author{Maxwill Lin}
\author{Augustine N. Mavor-Parker}
\author{Matthew James Sargent}

\affiliation{Vmax}

\abstract{\input{_abstract}}

\date{\today}
\correspondence{\emails{roger, augustine, matthew}}

\newcommand{\usepreprintmethodfigure}{}

\begin{document}
\maketitle

\input{_body}

\clearpage
\bibliography{curated_references}

\clearpage
\beginappendix
\input{appendix}

\end{document}

%% file: _numbers.tex
\newcommand{\baselineloraHumanEval}{76.8}
\newcommand{\baselineloraMBPP}{69.3}
\newcommand{\baselineloraLCB}{17.3}

\newcommand{\popHumanEvalMean}{78.7}
\newcommand{\popMBPPMean}{70.8}
\newcommand{\popLCBMean}{27.0}

%% file: _numbers_math.tex
\newcommand{\baselineloraAimetwofour}{6.7}

\newcommand{\baselineloraAmctwothree}{37.5}
\newcommand{\baselineloraMathfivehundred}{59.4}
\newcommand{\baselineloraMinerva}{19.1}

\newcommand{\baselineloraOlympiad}{27.3}

\newcommand{\popAimetwofourMean}{11.7}

\newcommand{\popAmctwothreeMean}{40.0}
\newcommand{\popMathfivehundredMean}{69.8}
\newcommand{\popMinervaMean}{27.0}

\newcommand{\popOlympiadMean}{33.9}

\newcommand{\baseloraMathMean}{34.1}

\newcommand{\popMathMeanMean}{38.2}

%% file: _abstract.tex
We introduce \textbf{PopuLoRA}, a population-based asymmetric self-play
framework for reinforcement learning with verifiable rewards (RLVR)
post-training of LLMs. Teachers and students are specialised LoRA
adapters on a shared frozen base: teachers propose problems, matched
students solve them under a programmatic verifier, and cross-evaluation
between sub-populations replaces the self-calibration that limits
single-agent self-play. A family of LoRA weight-space evolution
operators (mutations and crossovers that produce same-rank population
members in seconds) serves as the replacement step of a
population-based training loop at 7B scale. We instantiate PopuLoRA on
top of Absolute Zero Reasoner and compare it against a per-adapter compute-matched
single-agent baseline. Where the single agent self-calibrates to
generating easy problems it can reliably solve, the population enters a
co-evolutionary arms race: teachers produce increasingly complex
problems, student solve rates oscillate, and problem-space coverage
keeps expanding throughout training. Despite lower training-time
reward, the \emph{population mean} outperforms the baseline on three
code benchmarks (HumanEval+, MBPP+, LiveCodeBench) and seven math
benchmarks (AIME~24/25, AMC~23, MATH-500, Minerva, GSM8K,
OlympiadBench), and even the weakest member of the population beats
the baseline on aggregate.

%% file: _body.tex
\section{Introduction}
\label{sec:intro}

RL post-training has become a dominant regime for specialising large
language models, from RLHF~\citep{ouyang2022instructgpt} and DPO on
human preference data through self-play
fine-tuning~\citep{chen2024spin} and large-scale reinforcement
learning with verifiable rewards
(RLVR)~\citep{guo2025deepseek,lambert2024tulu}. The optimisation machinery
has matured, but the supply of \emph{problems} has not: most current
systems depend on a hand-curated task distribution whose scope,
difficulty, and coverage must be chosen in advance. The open question we
study in this paper is how to generate the \emph{curriculum itself}, the
stream of problems the policy is trained on, without relying on
human-authored datasets, using only a programmatic verifier as the
external signal.

The most direct approach lets a single model propose its own problems
and grade itself through the verifier, as in Absolute Zero
Reasoner~\citep{zhao2025absolute} or, more broadly, single-model
self-play~\citep{chen2024spin,languageselfplay2025}. The shared flaw is
that the same network generates problems and (implicitly, through its
solve rate or judgement) estimates their difficulty. In this paper we
provide empirical evidence that such single-agent self-generation
\emph{self-calibrates}: the proposer converges to generating problems it
can consistently produce in valid format and consistently solve, and the
training distribution collapses onto a narrow band long before the base
model's capability is exhausted. The fix is structural: make the judge a
\emph{different agent} from the proposer. A growing line of work
explores this asymmetry
\citep{spc2025,gasp2026,spice2025,soar2026,alive2026,triplay2026,ye2024eva,huang2025rzero},
but all are scalar in agent count ($\le$\,3 agents total), whether
the roles share parameters or train separate networks. We take the asymmetry further, building on the
teacher-proposes / student-solves structure introduced
by~\citet{sukhbaatar2018asym}, and replace the single agent with
co-evolving \emph{populations} of specialised \textbf{teachers} and
\textbf{students}. Teachers are rewarded for
producing problems that are hard for the particular student they face,
students are rewarded by the verifier, and matchmaking across the two
sub-populations turns difficulty into a population-level signal rather
than a self-estimate. Population dynamics add a second layer of
exploration on top of gradient updates: lineages diverge, members
specialise, and a population-based training
(PBT)~\citep{jaderberg2017population} replacement step recombines what
the gradient path has already discovered.

Running many independent full-parameter 7B models on a single node is
expensive and memory-constrained, particularly when each member must
support both rollout inference and gradient updates in a shared training
loop. We therefore instantiate every population member as a LoRA
adapter~\citep{hu2021lora} over a shared frozen base, which collapses
the population's memory footprint to the sum of its adapter weights
rather than a full copy of the base per member. Classical
PBT~\citep{jaderberg2017population} mutates members by copying one
agent's full weights onto another and perturbing; at 7B scale that copy
is itself expensive. Our second contribution is a set of
\textbf{LoRA weight-space evolution operators} (SVD-structured,
layer-selective, and component-masking mutations together with
DARE~\citep{yu2024dare}, TIES-inspired~\citep{yadav2023ties}, and
task-arithmetic~\citep{ilharco2023taskarithmetic} crossovers, in the
spirit of evolutionary model merging~\citep{akiba2024evolutionary})
that produce same-rank children in seconds without any retraining. They
serve as the replacement step of an \emph{online} PBT loop, making a
population-of-adapters regime that is inaccessible to prior LoRA
composition work~\citep{huang2023lorahub,buehler2024xlora,feng2024modelswarms,zhang2025genome}.
We evaluate in the Absolute Zero code-reasoning setting, with a
sandboxed Python executor as the verifier.

\begin{tcolorbox}[
  colback=white,
  colframe=black!50,
  boxrule=0.5pt,
  arc=1.5pt,
  left=6pt, right=6pt, top=5pt, bottom=5pt,
  boxsep=2pt,
]
{\normalfont\footnotesize\bfseries CONTRIBUTIONS\par}
\small
\vspace{4pt}
\begin{enumerate}[leftmargin=*, itemsep=3pt, topsep=0pt]
\item \textbf{Population-based asymmetric self-play for RLVR.} Prior
work~\citep{spc2025,gasp2026,spice2025,soar2026,alive2026,triplay2026,ye2024eva,huang2025rzero,languageselfplay2025}
is scalar in agent count ($\le$\,3 agents). PopuLoRA replaces these
with populations of teacher and student LoRA adapters coupled by
TrueSkill-weighted cross-evaluation.
\item \textbf{LoRA weight-space evolution operators} as the PBT
replacement step. Mutations and crossovers produce rank-matched
children in seconds; closest is evolutionary model
merging~\citep{akiba2024evolutionary}, applied \emph{offline} rather
than inside an online PBT loop.
\item \textbf{Empirical validation} on held-out code and math
benchmarks. The population outperforms a per-adapter compute-matched single-agent
baseline; diagnostics confirm it avoids the mode-collapse we observe
in the baseline.
\end{enumerate}
\end{tcolorbox}

\section{Background and Related Work}
\label{sec:background}

\paragraph{RLVR and self-generated curricula.}
RLVR~\citep{guo2025deepseek} replaces learned preference models with
programmatic checkers. AZR~\citep{zhao2025absolute} takes this to its
logical extreme: a single code LLM both proposes and solves its own
problems, rewarded only by a sandboxed executor, with no external
dataset. Adjacent methods
(STaR~\citep{zelikman2022star},
rStar-Math~\citep{rstarmath2025},
Self-Rewarding~\citep{yuan2024selfreward})
also train on self-generated data but rely on fixed problem sets or
learned reward signals; only AZR treats the model as both proposer and
verifier-checked solver. We evaluate PopuLoRA on the three
code-reasoning task types AZR defines: \texttt{code\_i}
(infer-input), \texttt{code\_o} (infer-output), and \texttt{code\_f}
(infer-function), all verified mechanically by the executor.

\paragraph{Self-play and asymmetric roles.}
Asymmetric self-play~\citep{sukhbaatar2018asym}, where one agent
proposes tasks and another solves them, is the structural ancestor
of PopuLoRA's teacher--student loop. In the unsupervised environment
design (UED) literature, PAIRED~\citep{dennis2020paired} formalises
regret-based adversarial curriculum generation,
POET~\citep{wang2019poet} co-evolves environments and agents, and
ACCEL~\citep{parkerholder2022accel} adds mutation operators on level
structure; emergent autocurricula arise in multi-agent
competition~\citep{baker2020emergent}. In the LLM setting, a growing
line of work separates proposer and solver:
SPIN~\citep{chen2024spin}, Language
Self-Play~\citep{languageselfplay2025},
SOAR~\citep{soar2026}, R-Zero~\citep{huang2025rzero},
ALIVE~\citep{alive2026}, TriPlay-RL~\citep{triplay2026}, and others
(Appendix~\ref{app:extrelated}). All are scalar in agent count
($\le$\,3 agents total). PopuLoRA differs in three ways: (i)~we
train \emph{populations} of teachers and students rather than a single
pair, (ii)~updates are joint and on-policy rather than alternating,
and (iii)~the difficulty signal comes from cross-evaluation across the
population rather than from a fixed target solve-rate band.

\paragraph{Population-based training and LoRA evolution.}
Classical PBT~\citep{jaderberg2017population} copies and perturbs full
agent weights; at 7B scale the per-member footprint and the
copy-and-perturb cost both become bottlenecks. Recent work lifts
evolution into adapter space (GENOME~\citep{zhang2025genome},
EGGROLL~\citep{eggroll2025}, ESSA~\citep{essa2025},
ES-at-Scale~\citep{esatscale2025}) but optimises against a
\emph{fixed} fitness function. PopuLoRA is orthogonal: evolution
serves as the replacement step of an RLVR self-play loop, where the
fitness signal is produced by the population itself through
cross-evaluation. We embed LoRA merge operators
(DARE~\citep{yu2024dare}, TIES-inspired~\citep{yadav2023ties},
task arithmetic~\citep{ilharco2023taskarithmetic}, plus SVD-structured
mutations) inside this \emph{online} loop, so children re-enter
gradient training immediately after recombination, unlike offline
merging~\citep{akiba2024evolutionary,feng2024modelswarms,huang2023lorahub}.
Further comparisons are in Appendix~\ref{app:extrelated}.

\section{\vmaxkeepcase{PopuLoRA}}
\label{sec:method}

\ifdefined\usepreprintmethodfigure
  \begin{figure}[t]
  \centering
  \includegraphics[width=0.75\linewidth]{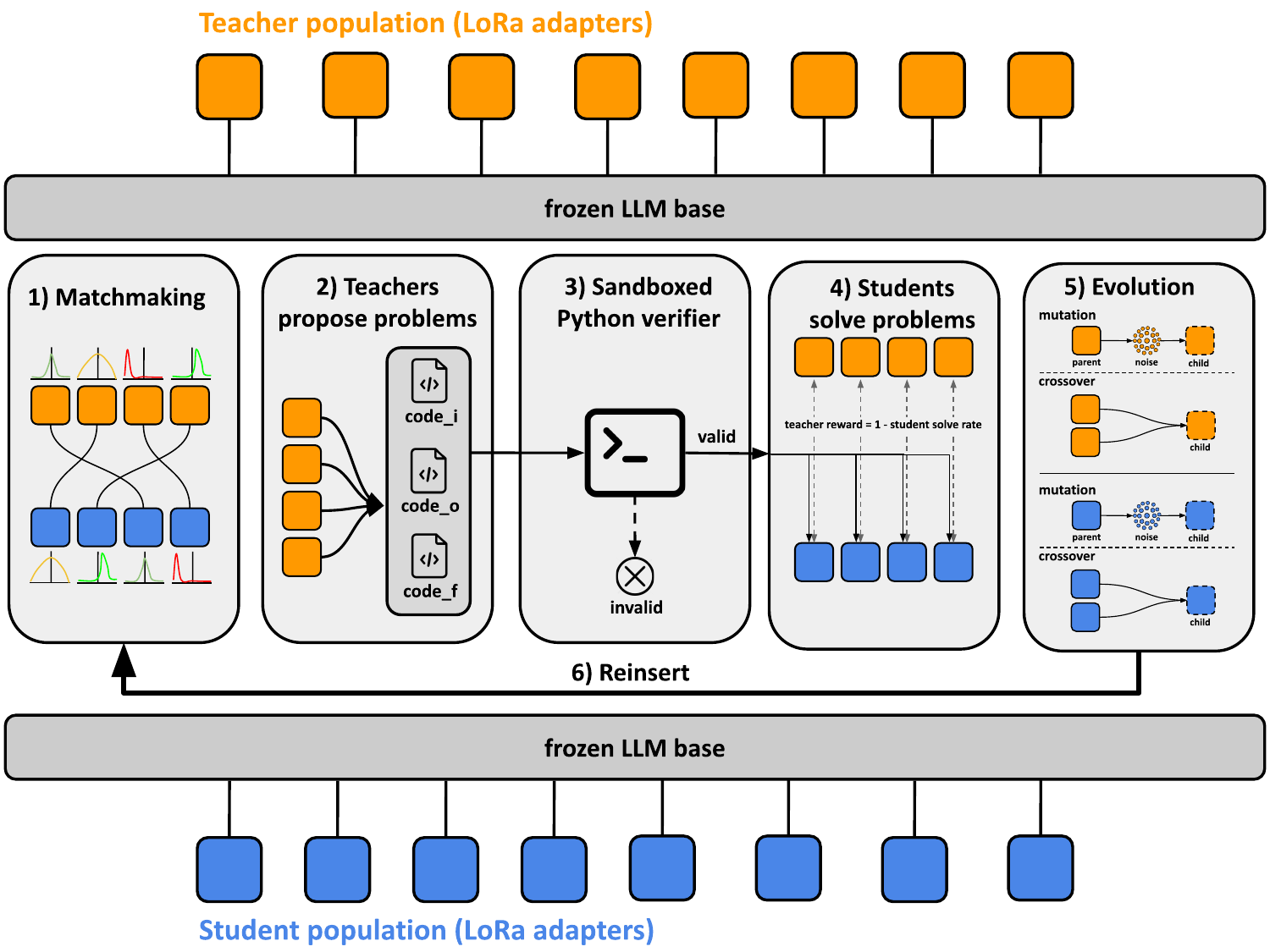}
  \caption{\textbf{One PopuLoRA iteration.} Matched teacher--student
  pairs generate and solve under a sandboxed verifier; the student's
  failure rate is the teacher's reward; every $k$ steps, LoRA evolution
  replaces the weakest members.}
  \label{fig:method}
  \end{figure}
\else
  \begin{wrapfigure}{r}{0.45\linewidth}
  \vspace{-14pt}
  \centering
  \includegraphics[width=\linewidth]{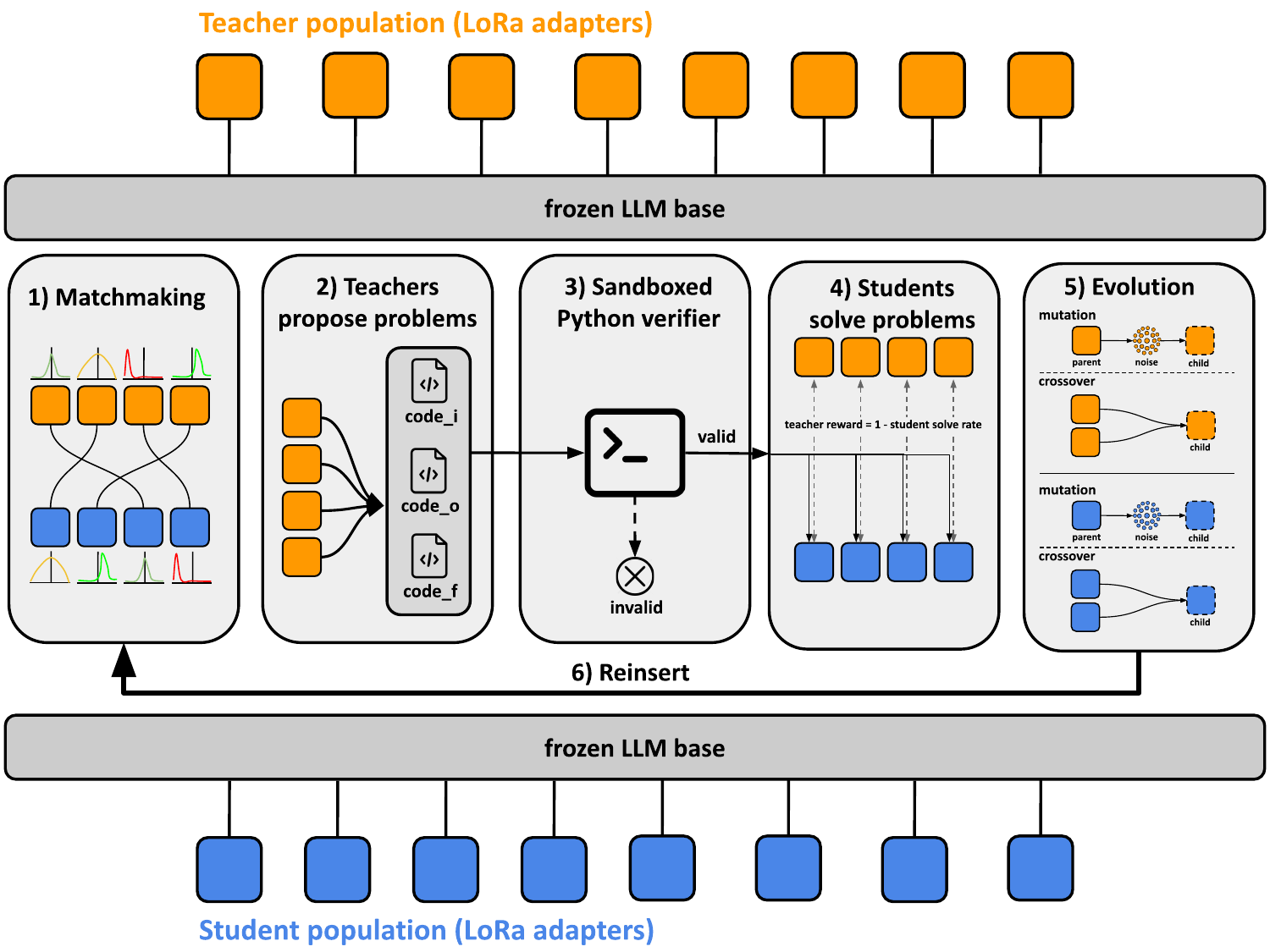}
  \caption{\textbf{One PopuLoRA iteration.} Matched teacher--student
  pairs generate and solve under a sandboxed verifier; the student's
  failure rate is the teacher's reward; every $k$ steps, LoRA evolution
  replaces the weakest members.}
  \label{fig:method}
  \vspace{-8pt}
  \end{wrapfigure}
\fi
PopuLoRA keeps AZR's core self-play loop but replaces the single agent
with a population of specialised LoRA adapters and replaces
self-calibrated difficulty with cross-evaluation between matched
members. Figure~\ref{fig:method} shows the resulting training step;
the rest of this section makes each component concrete.
Section~\ref{sec:method_objective} fixes the RLVR objective and reward
functions; Section~\ref{sec:method_arch} describes how the population
is parameterised; Section~\ref{sec:method_step} walks through one
training step; and Section~\ref{sec:method_evo} specifies the LoRA
weight-space evolution operators that serve as PopuLoRA's PBT
replacement step.

\subsection{Objective and reward}
\label{sec:method_objective}

Each population member optimises its policy by RLVR: a verifier $V$
(a sandboxed Python executor plus a format checker) emits a scalar
reward on every rollout, with no learned reward model in the loop.
For a student rollout $\tau$ on a problem $p$ proposed by its matched
teacher,
\begin{equation}
  R_{\text{stu}}(\tau) \;=\;
  \begin{cases}
    +1   & \text{if } V(\tau, p) = \text{correct,} \\
    -0.5 & \text{if } V(\tau, p) = \text{incorrect but well-formed,} \\
    -1   & \text{if the response fails format.}
  \end{cases}
  \label{eq:stu_reward}
\end{equation}
For a teacher-proposed problem $p$ subsequently attempted by student
$s$,
\begin{equation}
  R_{\text{tea}}(p) \;=\;
  \begin{cases}
    -1 & \text{if } p \text{ fails to parse, execute, or is non-deterministic,} \\
    0  & \text{if } \rho(t, s, p) = 0, \\
    1 - \rho(t, s, p) & \text{otherwise,}
  \end{cases}
  \label{eq:tea_reward}
\end{equation}
where $\rho(t, s, p)$ is the fraction of the student's rollout samples
that solve $p$. The zero-reward case when no student solves the problem
prevents teachers from being rewarded for generating impossible or
degenerate problems. The key structural change relative to single-agent
AZR is exactly in this equation: the teacher's reward depends on the
\emph{matched} student's failure rate, not on the proposer's own solve
rate, so difficulty is an inter-population quantity rather than a
self-estimate. Advantages are estimated with
REINFORCE++~\citep{hu2025reinforcepp,williams1992reinforce}-baseline
(per-prompt centring followed by global whitening across the batch),
in the critic-free GRPO~\citep{shao2024deepseekmath} family descended
from PPO~\citep{schulman2017ppo}, without a value network and without a KL
penalty to a reference model ($\beta_{\text{KL}}{=}0$). Every policy update pools all three AZR
problem types (\texttt{code\_i}, \texttt{code\_o}, \texttt{code\_f})
into a single mixed-type batch per member per step, matching the
single-agent baseline's pooling.

\subsection{Architecture}
\label{sec:method_arch}

The population consists of $N_T$ teacher and $N_S$ student LoRA
adapters attached to a single shared frozen code-LLM base. Every
adapter has the same rank $r$ and attaches to the same set of
projection matrices; only the adapter weights are updated, while the
base model remains frozen. This layout has two immediate consequences.
First, memory cost scales with the sum of adapter sizes rather than
the sum of full-parameter base copies, which is what makes
multi-adapter populations affordable on commodity hardware: a base
that is tens of gigabytes serves arbitrarily many adapters, each
costing only tens of megabytes. Second, the vLLM~\citep{kwon2023vllm} multi-LoRA
scheduler~\citep{sheng2023slora} can dispatch each request to the
correct adapter by tag inside a shared forward pass, so matched teachers and students generate and
solve in the same batch without any per-request base-model swap, and
swapping an adapter in or out of the rollout engine moves only the
adapter $\Delta W$, not the base. Concrete values of $N_T$, $N_S$,
$r$, and the base model used in our experiments are specified in
\S\ref{sec:exp_setup}.

\subsection{Training Step}
\label{sec:method_step}

Each training step proceeds in five phases.
\textbf{(1)~Matchmaking}: each teacher is paired with one student via
prioritised fictitious self-play
(PFSP)~\citep{vinyals2019alphastar} over
TrueSkill~\citep{trueskill2007} ratings, concentrating pairings on
informative near-balanced matchups
(Appendix~\ref{app:extrelated}). PopuLoRA adapts PFSP from
game-playing agents to problem-proposing teachers: opponents are
teachers generating problems rather than policies playing the same
game.
\textbf{(2)~Teacher generation}: each teacher proposes a batch of AZR
code problems (split across \texttt{code\_i}/\texttt{code\_o}/\texttt{code\_f}),
validated by the sandboxed executor.
\textbf{(3)~Student solve}: the matched student attempts all valid
problems under rollout; the executor produces a per-prompt binary
solve vector.
\textbf{(4)~Cross-evaluation and update}: the teacher's reward on
each valid prompt is $1 - \rho_{j_i}$ (Eq.~\ref{eq:tea_reward}),
where $\rho_{j_i}$ is the matched student's solve rate, replacing
AZR's self-estimation with an inter-population signal. All adapters
are updated in a single mixed batch.
\textbf{(5)~Evolution}: every $k$ steps, the bottom fraction $\gamma$
of each sub-population (ranked by TrueSkill lower-confidence bound) is
replaced by children produced via LoRA weight-space operators applied
to top-ranked parents (\S\ref{sec:method_evo}).

\subsection{LoRA Weight-Space Evolution}
\label{sec:method_evo}

The design goal is fast child creation: each operator has to emit a
same-rank LoRA adapter in seconds without any retraining, retain enough
parent knowledge that the child can re-enter training immediately, and
inject enough diversity to move the child away from its parents in
weight space. We implement two families of operators: mutations
(\textbf{M1--M4}) acting on a single parent and crossovers
(\textbf{X1--X4}) combining two parents. The eight operators described
here are the ones we use throughout training; the broader catalog and
the operators that failed retention tests are in
Appendix~\ref{app:operators}.

\paragraph{Mutations.}
Let $\Delta W = B A^\top \in \mathbb{R}^{d\times d}$ denote a single LoRA
layer's effective update. \textbf{M1 (SVD-structured)} takes the
singular-value decomposition $\Delta W = U \Sigma V^\top$, perturbs the
singular spectrum $\Sigma$ by a small multiplicative noise, and applies
a first-order Cayley rotation to $U, V$, preserving orthogonality while
moving the child off the parent's subspace. \textbf{M2 (layer-selective
Gaussian)} draws a random 33\,\% subset of the layer-module slots and
adds Gaussian noise to $A, B$ only within that subset; the remaining
two-thirds of the adapter are copied verbatim. \textbf{M3 (component
masking)} zeroes a random subset of the SVD components of $\Delta W$,
reducing effective rank and forcing the child to relearn masked
directions. \textbf{M4 (full Gaussian)} adds per-tensor adaptive
Gaussian noise to every $A, B$ factor, with the noise scale set by the
tensor's own running standard deviation.

\paragraph{Crossovers.}
Let $\Delta W^{(1)}, \Delta W^{(2)}$ be two parents. \textbf{X1 (DARE)}
applies the DARE drop-and-rescale recipe of~\citet{yu2024dare} to each
parent's delta then sums them. \textbf{X2 (layer-wise)} selects each
layer-module slot independently from either parent, giving a
layer-modular recombination reminiscent of TIES-style sign alignment
\citep{yadav2023ties}. \textbf{X3 (SVD subspace)} takes the top-$k$
singular components from one parent and fills the remaining $r-k$
components from the other, mixing parents within a common rank budget.
\textbf{X4 (extrapolative)} performs a task-arithmetic-style linear
combination \citep{ilharco2023taskarithmetic} with a coefficient greater
than one, extrapolating \emph{beyond} the convex hull of the parents
rather than interpolating between them. All eight operators run in
seconds on the adapter tensors alone, produce children at the same rank
as the parents, and are validated as retention-preserving in
\S\ref{sec:exp_retention}.

\section{Experiments}
\label{sec:experiments}

We evaluate PopuLoRA against a per-adapter compute-matched single-agent AZR baseline
on held-out code and math benchmarks, plus training-dynamics and
problem-complexity diagnostics, a population-dynamics analysis, a
LoRA-operator retention test, and a population-size ablation. All
comparisons share the same base model, reward scheme, and optimiser; the
only difference is whether training runs one self-calibrating agent or a
teacher--student population.

\subsection{Setup}
\label{sec:exp_setup}

\paragraph{Instantiation.}
The specific values below are the ones we use throughout our main
experiments; the method itself does not depend on any of these being
fixed to these values. We use a frozen
Qwen2.5-Coder-7B~\citep{hui2024qwen25coder} base, the same model
used by AZR, which enables a direct comparison against their publicly
released checkpoint. Every
population member is a rank-32 LoRA adapter with scaling $\alpha=64$
attached to the attention projections $\{W_q, W_k, W_v, W_o\}$. The
main population is $N_T=N_S=4$ (we sweep population size as an
ablation in \S\ref{sec:exp_ablations}). Training runs for 200 update
steps with AdamW at learning rate $5\times 10^{-5}$ and
REINFORCE++-baseline advantages (rollout $n=8$, temperature $T=1$,
no KL penalty). Each matchup generates a batch of 72 prompts split
equally across \texttt{code\_i}, \texttt{code\_o}, \texttt{code\_f};
the student solves the same split. Evolution runs every $k=10$ steps
on the bottom fraction $\gamma=0.25$ of each sub-population ranked by
TrueSkill lower-confidence bound (ablated in
\S\ref{sec:exp_ablations}). Full hyperparameters are listed in
Appendix~\ref{app:hyperparams}.

\paragraph{Evaluation.}
At evaluation time we merge each final adapter into the frozen base
with PEFT and report greedy pass@1 on (i)~code benchmarks
HumanEval+~\citep{chen2021humaneval,liu2023evalplus},
MBPP+~\citep{austin2021mbpp,liu2023evalplus}, and
LiveCodeBench~v5~\citep{jain2024livecodebench}, and (ii)~math
benchmarks AIME~24/25, AMC~23,
MATH-500~\citep{hendrycks2021math,lightman2023verify},
Minerva~\citep{lewkowycz2022minerva},
GSM8K~\citep{cobbe2021gsm8k}, and
OlympiadBench~\citep{he2024olympiadbench}. For the
population we report the population mean as the headline result, and
include the per-benchmark argmax across adapters as a test-set-selected
upper bound (not a deployable selection rule, since the choice of
adapter depends on the benchmark scores); per-adapter breakdowns are in
Appendix~\ref{sec:app_full_ft_comparison}.

\subsection{Downstream Benchmarks}
\label{sec:exp_downstream}

\begin{figure}[ht!]
\centering
\includegraphics[width=\linewidth]{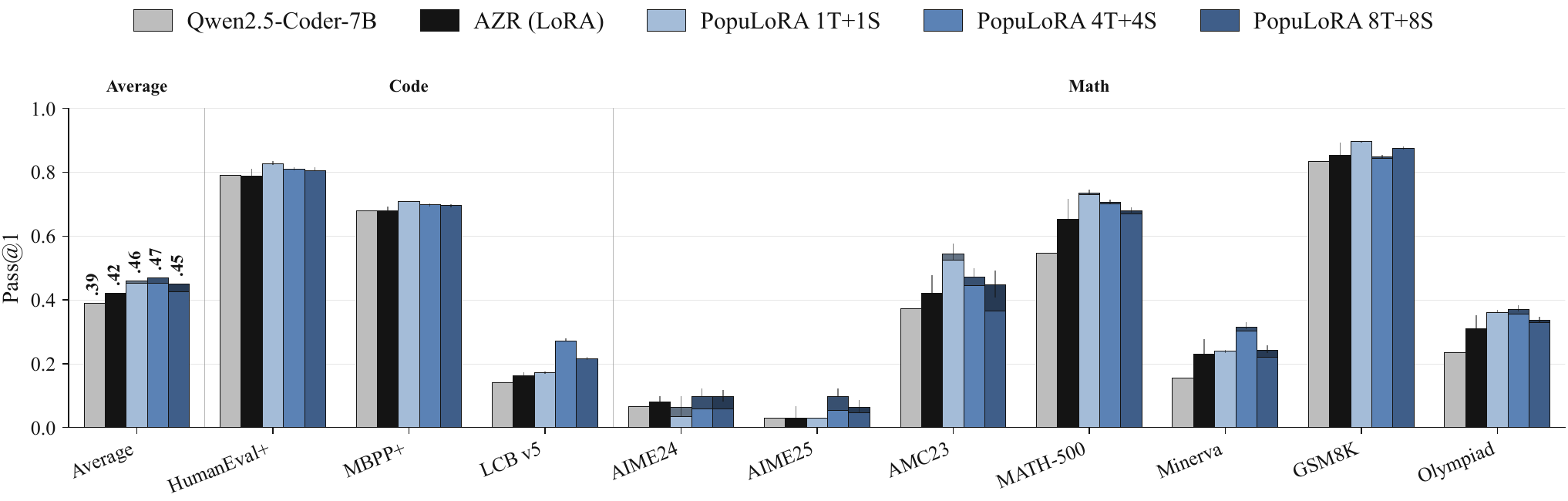}
\caption{\textbf{Downstream pass@1.} Baseline AZR (LoRA) is
per-adapter compute-matched. The 8T+8S population is evaluated at 100
gradient steps. For each population size, the lighter bar (the
headline number) shows the mean across all adapters; the darker cap
shows the per-benchmark argmax across adapters --- a test-set-selected
upper bound, not a deployable result, since the choice of adapter
depends on the benchmark scores. Error bars show $\pm$1 std across
adapters. Full per-adapter breakdown in
Appendix~\ref{sec:app_full_ft_comparison}.}
\label{fig:downstream}
\end{figure}

Figure~\ref{fig:downstream} reports greedy pass@1 across three code and
seven math benchmarks, comparing the population's mean and best teacher and best
student against the per-adapter compute-matched baseline. The 8T+8S
rows use their available 100-gradient-step checkpoint.

\paragraph{Code.}
The population mean is at or above the per-adapter compute-matched
baseline on every code benchmark (Figure~\ref{fig:downstream}):
\textbf{\popHumanEvalMean} vs.\ baseline LoRA \baselineloraHumanEval{}
on HumanEval+, \textbf{\popMBPPMean} vs.\ \baselineloraMBPP{} on
MBPP+, and \textbf{\popLCBMean} vs.\ \baselineloraLCB{} on
LiveCodeBench. The largest gap sits on LiveCodeBench, which contains
recent competitive-programming problems from online contests and is
therefore structurally different from the self-generated code triples
AZR trains on.

\paragraph{Math.}
Despite AZR providing no direct math supervision (its verifier is a
Python executor on code triples, not a math grader), PopuLoRA shows
consistent out-of-domain gains. The population mean improves on the
baseline LoRA on every math benchmark we measured:
\textbf{\popAimetwofourMean} vs.\ \baselineloraAimetwofour{} on
AIME~24, \textbf{\popAmctwothreeMean} vs.\ \baselineloraAmctwothree{}
on AMC~23, \textbf{\popMathfivehundredMean} vs.\
\baselineloraMathfivehundred{} on MATH-500, \textbf{\popMinervaMean}
vs.\ \baselineloraMinerva{} on Minerva, and
\textbf{\popOlympiadMean} vs.\ \baselineloraOlympiad{} on
OlympiadBench, with an average math gain of
\textbf{\popMathMeanMean} vs.\ \baseloraMathMean{}. Both arms improve
over the frozen base, but the population's margin is larger on every
math benchmark, with the biggest absolute gaps on the competition-level
benchmarks (AIME, OlympiadBench) where the baseline barely moves from
the base model. Even the weakest member of the 4T+4S population beats
the baseline on aggregate (Table~\ref{tab:downstream}), confirming that
co-evolution lifts the entire population rather than concentrating
gains in a few specialists. Per-benchmark argmax over the four adapters
of each role (a test-set-selected upper bound, not a deployable
selection rule) lifts the math gains further, particularly on AIME and
AMC, where the population was never directly trained: see
Table~\ref{tab:downstream} for the full per-adapter breakdown.

\subsection{Training Dynamics: Self-Calibration vs.\ Co-Adaptation}
\label{sec:exp_dynamics}

\begin{figure}[ht!]
\centering
\includegraphics[width=\linewidth]{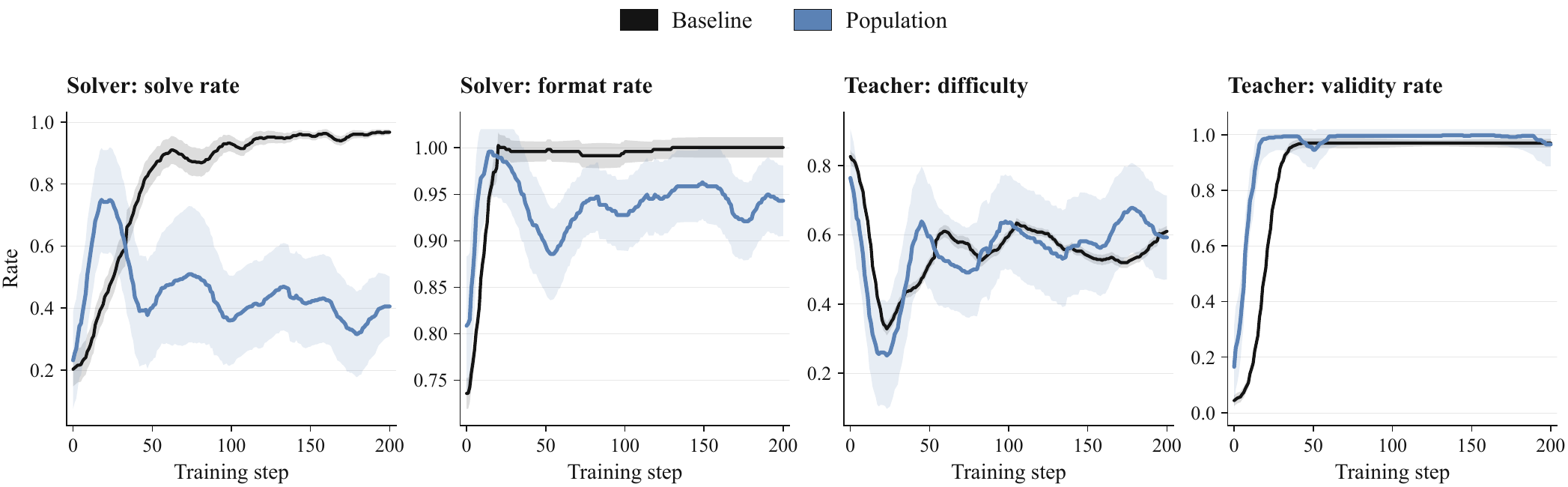}
\caption{\textbf{Training dynamics.} Left two panels: solver
(solve rate, format rate). Right two panels: teacher (problem
difficulty $= 1 - \text{solve rate}$, validity rate). Baseline in
black, population mean in blue with per-member spread. Per-type
breakdown in Appendix~\ref{app:pertype}.}
\label{fig:dynamics}
\end{figure}

The training dynamics of the baseline and the population tell two
qualitatively different stories (Figure~\ref{fig:dynamics}).

In the single-agent baseline, the solver's solve rate rises
monotonically and plateaus within the first 200 steps. A near-perfect
solve rate is not the right signal here: it means the teacher is unable
to generate problems hard enough to make the student fail. The single
agent simultaneously learns to produce valid problems in the correct format
\emph{and} to solve everything it generates. There is no increasing
complexity, no co-adaptation, no pressure to improve further: the
system has found a stable fixed point where both roles are satisfied
with minimal effort.
Notably, the teacher-difficulty curves can look superficially similar
between baseline and population, but they mean different things. In the
baseline, the teacher is rewarded against its own solver's solve rate
--- difficulty is a self-estimate by the same network that proposes
the problem; in the population, student failures directly reflect a
separate, matched student's solve rate, so difficulty is an
inter-population signal.

The population's dynamics are strikingly different. Student solve rates
oscillate throughout training rather than monotonically rising. This
pattern has a natural explanation: as teachers co-adapt and generate
harder problems, students start failing; once students catch up,
teachers are pushed to produce yet harder problems, and the cycle
repeats. This phasic dynamic is the signature of a genuine
co-evolutionary arms race rather than a self-calibrating fixed point.

Crucially, the population's seemingly lower training-time solve rate is
a sign of strength. When we evaluate the end-of-training checkpoints
on held-out benchmarks (Figure~\ref{fig:downstream}), they consistently
outperform the baseline whose training curve looked near-perfect. The
population's teachers kept pushing difficulty upward, which forced the
students to develop capabilities the baseline never needed.

The per-type breakdown (Appendix~\ref{app:pertype},
Figure~\ref{fig:dynamics_per_type}) reveals that the oscillation runs at
different frequencies across task types: \texttt{code\_o} cycles rapidly
between teacher- and student-dominated phases, while \texttt{code\_i}
and \texttt{code\_f} oscillate more slowly, consistent with these tasks
requiring deeper program understanding to sustain difficulty pressure.
The baseline's policy entropy collapses to near zero while the
population's teachers maintain non-trivial entropy throughout training
(Appendix~\ref{app:diagnostics}), and the population's response length
grows to ${\sim}1000$ tokens versus the baseline's ${\sim}250$
(Appendix~\ref{app:response_length}), consistent with more elaborate
reasoning~\citep{guo2025deepseek}.

\subsection{Problem Complexity: Collapse vs.\ Growth}
\label{sec:exp_modecollapse}

\begin{figure}[ht!]
\centering
\includegraphics[width=\linewidth]{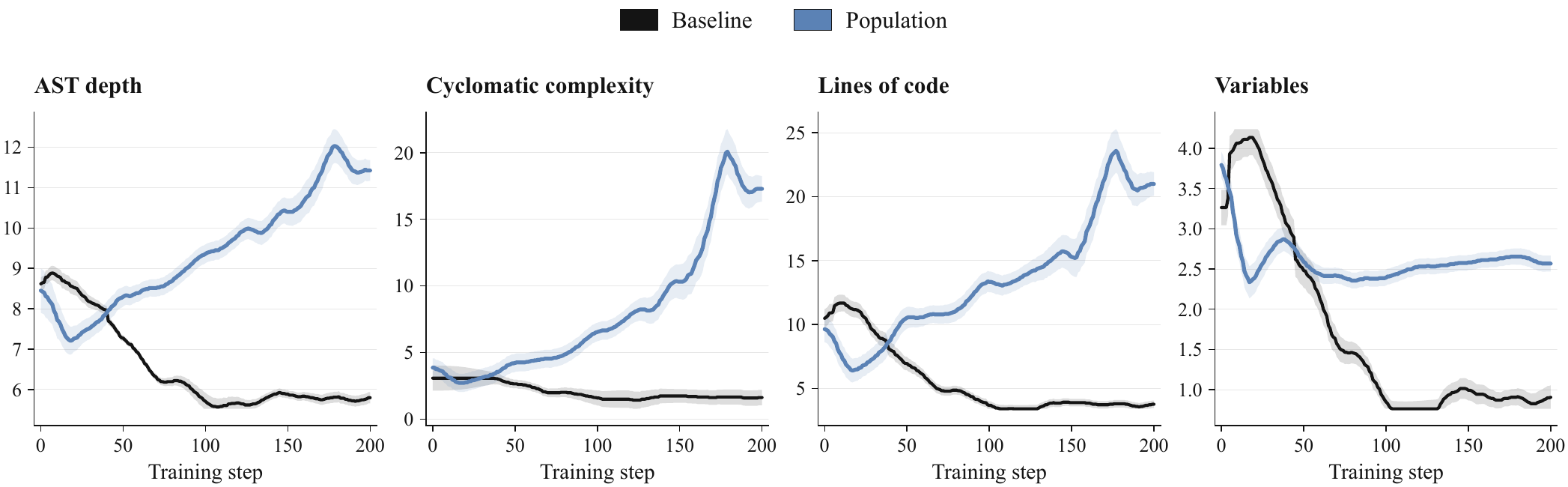}
\caption{\textbf{Program complexity over training.} Baseline (black)
trends downward on all four axes; population (blue) trends upward.
Coverage analysis in Appendix~\ref{app:coverage}.}
\label{fig:modecollapse}
\end{figure}

The training-dynamics story is corroborated by direct measurements of
problem complexity (Figure~\ref{fig:modecollapse}). We track four
structural metrics of teacher-generated programs over training: AST
depth, cyclomatic complexity, lines of code, and variable count
(definitions in App.~\ref{app:coverage}).

The difference is clear. In every panel the baseline curves trend
\emph{downward}: the single-agent teacher learns to produce
progressively simpler programs along every axis, converging on the
simplest programs it can consistently generate in valid format and solve.
The population teachers show the opposite trajectory: all four
complexity metrics rise throughout training. Cross-evaluation rewards
teachers for problems that are hard for the matched student, which
creates sustained upward pressure on problem difficulty rather than
the downward drift of self-calibration.

Problem-space coverage tells the same story: we tile the structural
feature space with a CVT archive~\citep{vassiliades2018cvt} of
4\,096 cells and track the fraction filled over training
(Appendix~\ref{app:coverage}). Baseline coverage plateaus early; the
population keeps expanding through 200 steps, generating increasingly
diverse problems alongside increasingly complex ones. By step~100 the
baseline has collapsed to programs as trivial as
\texttt{return number * 3} (Appendix~\ref{app:samples}).

\subsection{Population Dynamics: Arms Race and Specialisation}
\label{sec:exp_popdyn}

\begin{figure}[ht!]
\centering
\includegraphics[width=\linewidth]{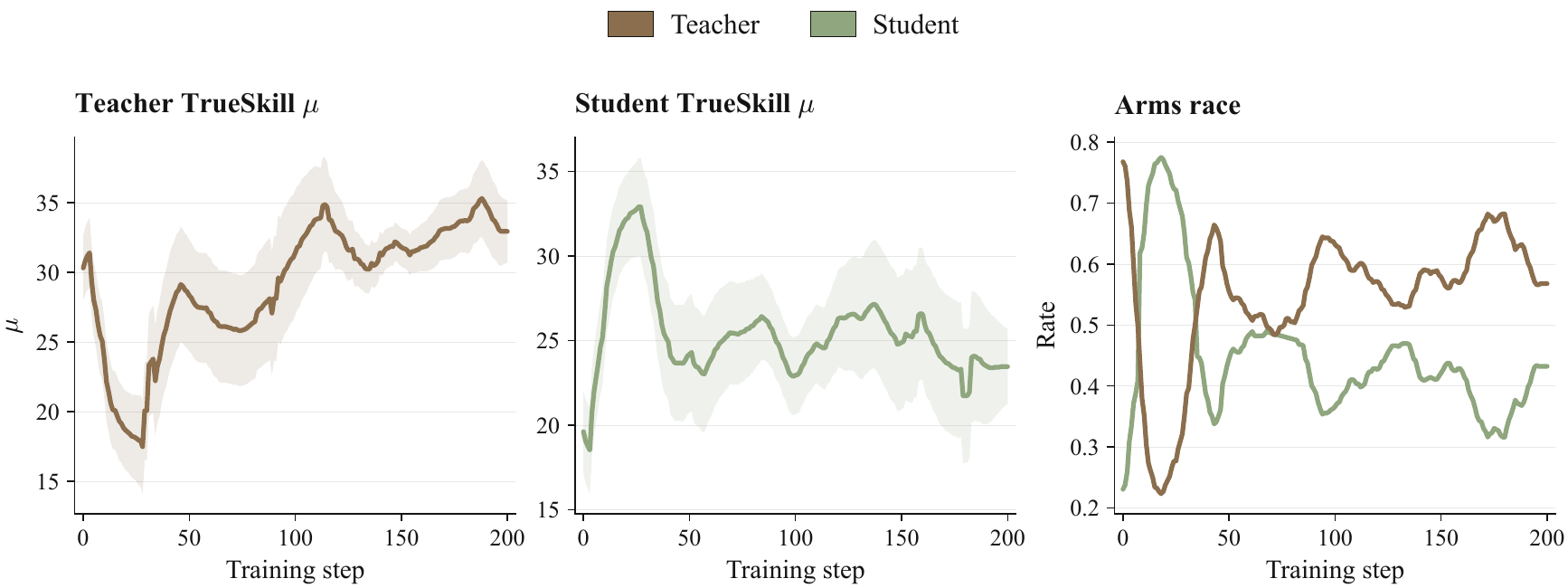}
\caption{\textbf{TrueSkill $\mu$ and arms race.} Left/centre:
per-adapter ratings (light) and role mean (bold). Right: matchup
outcome from student (blue) vs.\ teacher (orange) perspective; the
lead alternates throughout training.}
\label{fig:trueskill}
\end{figure}

The TrueSkill ratings provide a direct view of the co-evolutionary
dynamics (Figure~\ref{fig:trueskill}). Three features stand out.

Individual adapters differentiate from the population mean as training
progresses. Early on, all members cluster near the prior $\mu{=}25$;
by mid-training, distinct high and low performers have emerged in both
sub-populations, indicating that the population dynamics produce
genuine specialisation rather than homogeneous copies.

The right panel reveals an oscillating arms race between the two roles.
There are periods where teachers dominate --- their problems are too
hard for the matched students --- but students eventually catch up,
and the cycle repeats. This alternating lead is a characteristic of
adversarial co-evolution: neither role is able to settle into a fixed
strategy because the other keeps adapting. Per-student solve-rate
profiles against different teachers are non-uniform
(Appendix~\ref{app:crosseval}), confirming that specialisation extends
to the matchup level.

\subsection{LoRA Operator Retention}
\label{sec:exp_retention}

\begin{figure}[H]
\centering
\includegraphics[width=\linewidth]{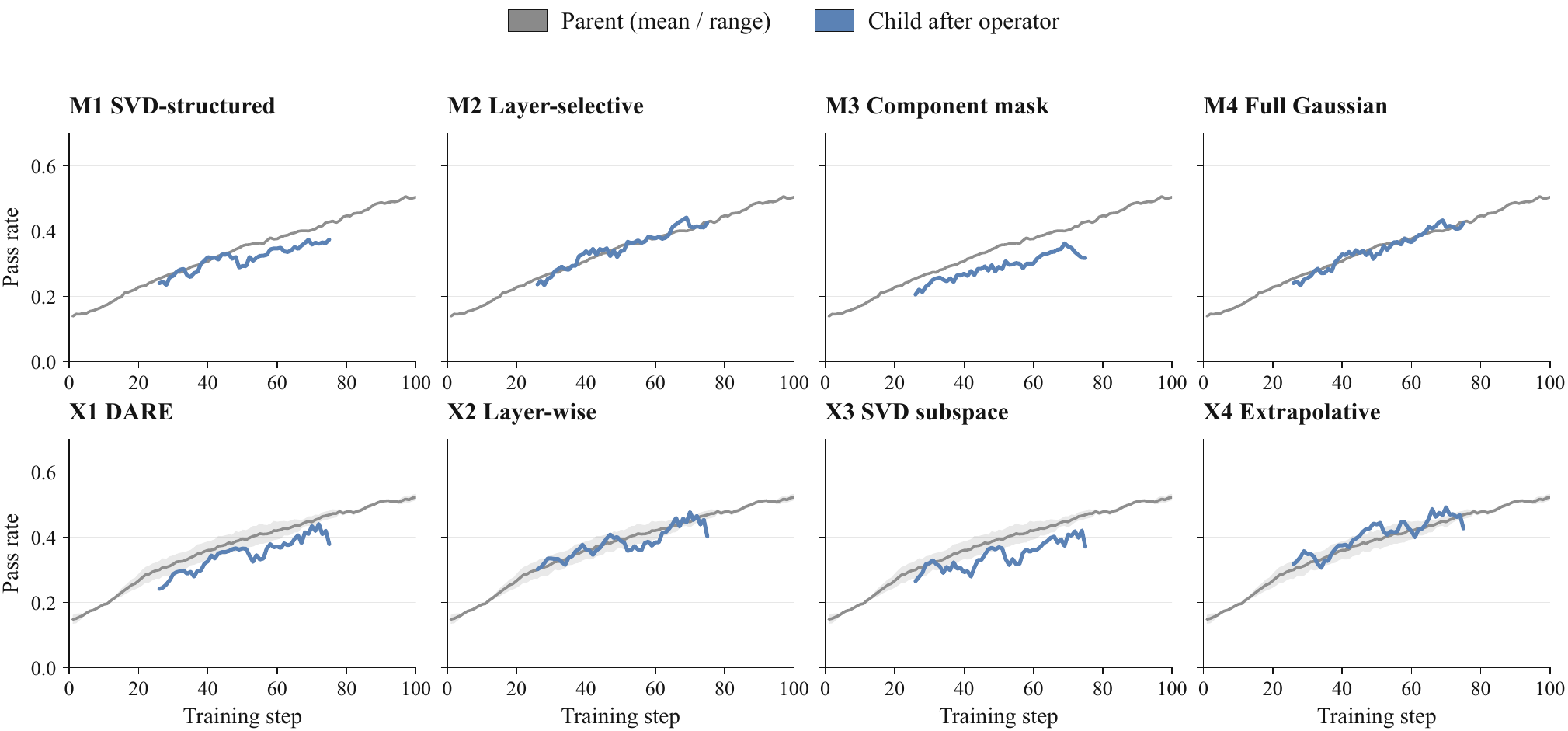}
\caption{\textbf{LoRA operator retention (snapshot step 25).} Top:
mutations (parent in grey). Bottom: crossovers (two parents in grey;
trained on different task types). All children recover to
near-parent performance within ${\sim}20$ steps. Full operator grid
in Appendix~\ref{app:loraops}.}
\label{fig:retention}
\end{figure}

All eight operators that ship in the live population pass the retention
test (Figure~\ref{fig:retention}): every child recovers to parent-level
reward within 10--20 update steps after being re-injected into training.

\FloatBarrier
\FloatBarrier

The mutation results (top row) confirm that perturbed children start
close to their parent and resume gradient updates without resetting to
the frozen base, validating the operators as a legitimate PBT
replacement step. The crossover results (bottom row) combine parents
trained on different AZR task types (e.g.\ one specialised on induction
and the other on output prediction). The crossover child retains
performance on \emph{both} parents' tasks, demonstrating that
weight-space recombination can compose complementary specialisations
into a single adapter. See Appendix~\ref{app:loraops} for the full
operator grid across all snapshot steps.

\subsection{Population Size Ablation}
\label{sec:exp_ablations}

\begin{figure}[H]
\centering
\includegraphics[width=\linewidth]{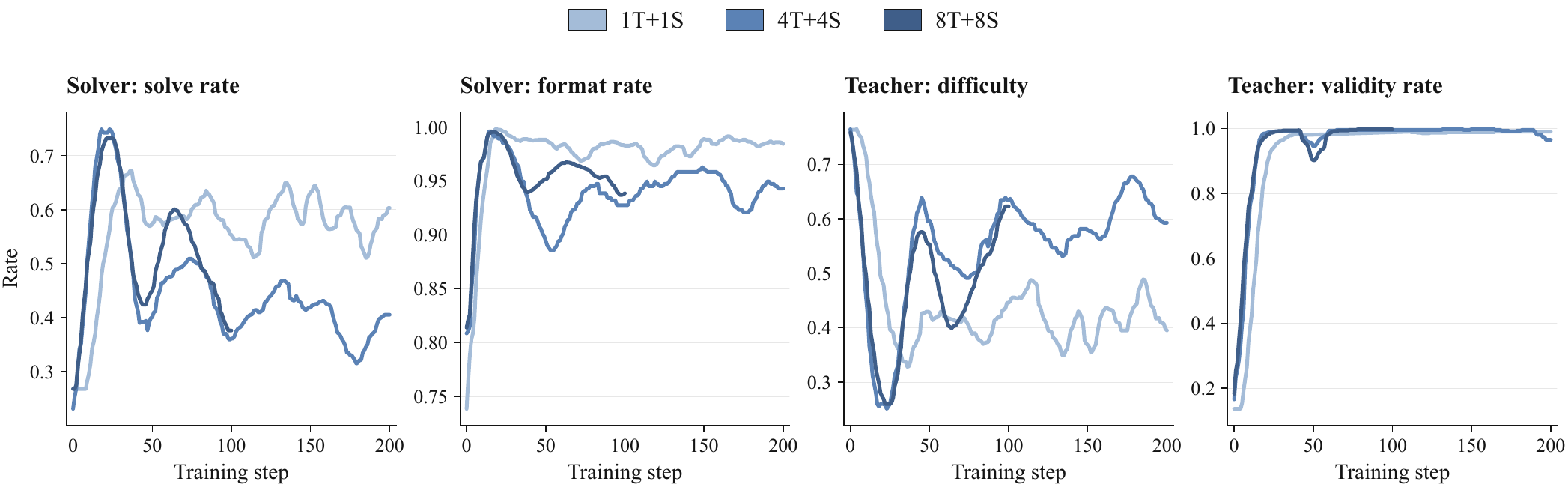}
\caption{\textbf{Population size ablation.} Even a single
teacher--student pair (1T+1S) avoids the baseline's mode collapse.
Co-evolutionary oscillations become more pronounced at 4T+4S and
8T+8S. The 8T+8S run shown here stops at 100 gradient steps.}
\label{fig:ablation_size}
\end{figure}

The population size ablation (Figure~\ref{fig:ablation_size}) reveals
that the co-evolutionary dynamics we observe are not simply a
consequence of having two roles --- they require a population.

\FloatBarrier
\FloatBarrier

Even at the smallest population size, 1T+1S, decoupling the teacher
and student into separate adapters is enough to avoid the baseline's
mode collapse: the solver reward does not plateau at near-perfect
levels, and downstream evaluation
(Table~\ref{tab:downstream}) shows that 1T+1S already outperforms
the single-agent baseline on most benchmarks. The oscillation pattern
is less pronounced than in larger populations, but the structural
separation of proposer and solver prevents the self-calibrating fixed
point that limits single-agent training.

At 4T+4S the co-evolutionary oscillations become clearly visible and
the reward dynamics show the phasic arms race described above. This
is the configuration we use for the main comparison. At 8T+8S the
oscillations are more pronounced, though the downstream gains depend
on the benchmark (Table~\ref{tab:downstream}); both the ablation trace
and downstream evaluation for 8T+8S use the 100-gradient-step checkpoint.

\section{Discussion and Conclusion}
\label{sec:discussion}

PopuLoRA replaces single-agent self-calibration with co-evolving
teacher--student LoRA populations on a shared frozen base. Where the
single-agent baseline converges to a fixed point of easy problems and
near-perfect reward, the population sustains an arms race that produces
increasingly complex problems and stronger downstream performance on
every code and math benchmark we measured. Even the weakest member of
the 4T+4S population outperforms the baseline on aggregate, indicating
that co-evolution lifts the entire population rather than concentrating
gains in a few specialists (Table~\ref{tab:downstream}). Even a
minimal 1T+1S configuration, which decouples proposer and solver
without population dynamics, already improves over the baseline,
confirming that structural asymmetry is the primary driver.

\paragraph{Compute cost.}
Per-adapter compute matches the baseline: same tokens, rollouts, and
updates per step. Total work scales with adapter count, but wall-clock
scales sub-linearly because vLLM multi-LoRA batching shares the frozen
base forward pass~\citep{sheng2023slora}. On the same
1$\times$8$\times$H100 node, 4T+4S trains $8{\times}$ more adapters for
only $1.31{\times}$ the baseline wall-clock
(Appendix~\ref{app:hyperparams}, Table~\ref{tab:compute}).

\paragraph{What LoRA evolution does not replace.}
Evolution operators recombine and perturb existing adapters; new
behaviour still comes from policy-gradient updates after injection.

\paragraph{Limitations.}
We fix the LoRA rank to 32 across all population members; heterogeneous
ranks could further diversify the population but we have not explored
this axis. All experiments use a single base model
(Qwen2.5-Coder-7B-Instruct) and a single verifier domain (sandboxed
Python execution), so it remains open whether the co-evolutionary
dynamics transfer to other base scales or to domains with weaker
verifiers. Code will be released upon publication.

%% file: appendix.tex
% paper/appendix.tex

\section{Extended Related Work}
\label{app:extrelated}

\paragraph{Further asymmetric self-play methods.}
We expand here on the asymmetric self-play methods only briefly
summarised in the main paper.
SPIRAL~\citep{liu2025spiral} extends shared-policy self-play to
two-player zero-sum games (TicTacToe, Kuhn Poker, Simple Negotiation)
with role-conditioned advantages: it is genuinely symmetric in
the sense that both players sample from the same policy.
SPC~\citep{spc2025} trains a step-level process-reward critic
against a ``sneaky generator'' that injects subtle reasoning errors;
this is a process-reward-model training recipe rather than a
problem-solving self-play setup.
GASP~\citep{gasp2026} grounds a coding teacher on real hard
``goalpost'' problems the student cannot yet solve and asks the
teacher to emit easier (``lemma'') and harder (``lift'') variants
bridging the gap: the distinctive mechanism is the use of real
problems as anchor points, which neither AZR nor PopuLoRA relies on.
SPICE~\citep{spice2025} gives its Challenger document access that
the Reasoner lacks; document asymmetry drives the adversarial
curriculum.
eva~\citep{ye2024eva} targets \emph{RLHF} (not RLVR) by evolving
prompt distributions with a regret-based estimate-sample-evolve
procedure over a shared policy; the reward is a learned preference
model, not a programmatic verifier.

\paragraph{Further LoRA-space and evolutionary methods.}
LoRAHub~\citep{huang2023lorahub} black-box-searches mixture
coefficients over a library of pre-trained LoRA experts at test
time, without any weight-space gradient. X-LoRA~\citep{buehler2024xlora}
token-level-gates between frozen LoRAs at inference, producing
dynamic mixtures without touching the adapter weights.
Heterogeneous Swarms~\citep{feng2025heterogeneous} jointly
optimises model roles (as DAG adjacency) and weights under PSO.
Promptbreeder~\citep{promptbreeder2023} evolves text artifacts
(prompts) over a fixed LLM with an analogous fitness-and-mutation
structure, but does not touch model weights.
ESSA~\citep{essa2025} restricts ES to low-rank attention adapters
compressed via SVD and runs entirely in low-precision inference,
targeting post-SFT alignment.
ES-at-Scale~\citep{esatscale2025} (Cognizant) reports the first
successful full-parameter ES fine-tuning of multi-billion-parameter
LLMs, beating PPO/GRPO on sparse-reward reasoning tasks such as
Countdown.
Spherical linear interpolation (SLERP) is a standard alternative to
linear midpoints when the two merged vectors are near anti-parallel.
None of these operates inside an online PBT loop over a co-evolving
adversarial partner, which is the regime PopuLoRA targets.

\paragraph{TrueSkill and matchmaking details.}
We use TrueSkill~\citep{trueskill2007}, a Bayesian skill-rating
system designed for head-to-head matches, to produce a per-adapter
rating for each role at every step. Each adapter carries a pair
$(\mu, \sigma^2)$: after a matchup where teacher $t$'s problems are
solved at rate $\rho$ by student $s$, we record the matchup as a win
for whichever role came out above its conditional expectation (the
teacher wins if $\rho$ is small enough that its problem was too hard
for the student on average, and vice versa), and update
$(\mu, \sigma)$ for both adapters under the standard TrueSkill
Bayesian update. Ratings then drive two things: (i)
\emph{matchmaking}: we sample the next opponent by prioritised
fictitious self-play
(PFSP)~\citep{vinyals2019alphastar,heinrich2016nfsp}, weighting draw
probability by the TrueSkill-predicted win rate, so pairings concentrate on informative
near-balanced matchups rather than mismatches; and (ii)
\emph{culling}: at evolution time we rank each sub-population by the
TrueSkill \emph{lower-confidence bound} $\mu - k\sigma$ and replace
the bottom fraction, which penalises low-$\mu$ adapters but also
high-$\sigma$ (under-sampled) ones.

\section{Hyperparameters}
\label{app:hyperparams}

Table~\ref{tab:hyperparams} lists the full configuration used for every
population member and the single-agent baseline. All values are held
identical across arms unless a row explicitly splits them.

\begin{table}[h]
\centering
\begin{tabular}{ll}
\toprule
Setting & Value \\
\midrule
Base model                   & Qwen2.5-Coder-7B \\
LoRA rank $r$                & 32 \\
LoRA $\alpha$                & 64 \\
LoRA targets                 & q\_proj, k\_proj, v\_proj, o\_proj \\
Optimizer                    & AdamW \\
Learning rate                & $5\times 10^{-5}$ \\
Mini-batch size              & 64 \\
Micro-batch per GPU          & 4 (pop\_4t4s) / 2 (pop\_16t16s) \\
Rollout $n$                  & 8 \\
Temperature                  & 1.0 \\
Advantage estimator          & REINFORCE++-baseline \\
KL regularisation            & disabled \\
Max prompt length            & 6144 \\
Max response length          & 8096 \\
Per-step prompts (per role)  & 24 + 24 + 24 across code\_i / code\_o / code\_f \\
Evolution interval           & 10 steps \\
Evolution cull fraction      & 25\% \\
Total training steps         & 200 \\
Hardware                     & 1 $\times$ 8$\times$H100 (80\,GB) \\
\bottomrule
\end{tabular}
\caption{\textbf{Full PopuLoRA hyperparameter configuration.} All
rows are shared between the baseline and population runs unless
otherwise noted.}
\label{tab:hyperparams}
\end{table}

\paragraph{Per-adapter compute is identical; wall-clock scales
sub-linearly thanks to multi-LoRA batching.}
By design, no hyperparameter is tuned per arm: every population
member is configured identically to the single-agent baseline: same
base model, rank, learning rate, rollout $n$, per-prompt budget,
advantage estimator, and training horizon. At each training step the
baseline produces one matchup's worth of rollouts and one
single-adapter gradient update, while the population produces $N_T$
matchups and updates all $N_T+N_S$ adapters in a single mixed
mega-batch. Consequently,
\begin{align*}
  \text{rollouts / step (pop)}     &= (N_T + N_S) \times \text{rollouts / step (baseline)}, \\
  \text{grad updates / step (pop)} &= (N_T + N_S) \times \text{grad updates / step (baseline)}.
\end{align*}
Every adapter in the population therefore sees the same number of
tokens, the same rollout distribution size, and the same number of
policy updates per step as the baseline agent does. The only
step-scoped factor that changes between arms is the number of
adapters training in parallel, so observed gaps are attributable to
population dynamics rather than extra compute per adapter, a larger
rollout budget per prompt, or a richer optimiser configuration.

\paragraph{Compute accounting.}
Table~\ref{tab:compute} reports the actual wall-clock cost of each
configuration on the same hardware (1$\times$8$\times$H100 node).
The vLLM multi-LoRA scheduler~\citep{sheng2023slora} batches all
adapters through a shared base-model forward pass, so rollout time
is dominated by the base model rather than the adapter count. This
gives dramatic sub-linear wall-clock scaling: the 4T+4S population
trains $8{\times}$ more adapters for only $1.31{\times}$ the
wall-clock, yielding a $6.1{\times}$ throughput gain in
adapter-steps per hour. All configurations run on identical hardware
with no additional nodes.

\begin{table}[h]
\centering
\begin{tabular}{lrrrr}
\toprule
Configuration & Adapters & Adapter-steps & Wall-clock (h) & Ratio vs.\ baseline \\
\midrule
Baseline AZR (LoRA) & 1  & 200   & 61   & $1.00{\times}$ \\
PopuLoRA 4T+4S      & 8  & 1\,600 & 80   & $1.31{\times}$ \\
PopuLoRA 8T+8S      & 16 & 3\,200 & 110  & $1.81{\times}$ \\
\bottomrule
\end{tabular}
\caption{\textbf{Compute accounting at 200 training steps.} All runs
use the same 1$\times$8$\times$H100 (80\,GB) node. Wall-clock is
the median per-step time (excluding the first 5 warm-up steps)
extrapolated to 200 steps. The 4T+4S population trains
$8{\times}$ more adapters at only $1.31{\times}$ the
baseline wall-clock cost.}
\label{tab:compute}
\end{table}

\paragraph{Scaling ceiling.}
\label{app:failures}
Scaling the population on a single node is bounded by two resources
that grow with the adapter count: per-adapter activation memory on
the backward pass, and collective-op timeouts when heterogeneous
per-rank work (dominated by long-running sandbox validations on
some ranks) blocks the allreduce. At the population sizes we
report, both are manageable with tighter micro-batching and a raised
NCCL watchdog; beyond that, the scaling headroom is set by the
interaction between per-adapter backward memory, the vLLM KV-cache
reservation for rollouts, and rank-heterogeneous validation time.
Sharding the base across nodes or offloading validation to a
dedicated worker pool is the natural next step.

\section{Full LoRA Operator Catalog}
\label{app:operators}

We catalogue all 17 operators implemented in the \texttt{experiments/lora\_ops/}
benchmark: six mutations, nine crossovers, and two identity controls. Every
operator consumes one or two rank-$r$ LoRA state dicts and emits a
rank-matched child in seconds, without any retraining. For mutations we
write $\Delta W = B A^\top$ for a single LoRA module, where
$A \in \mathbb{R}^{r \times d}$ and $B \in \mathbb{R}^{d \times r}$; for
crossovers we mark parents with superscripts
$\Delta W^{(1)}, \Delta W^{(2)}$. The effective-delta SVD used by several
operators is computed via a double-QR on the $r \times r$ core (see
\texttt{\_efficient\_svd\_of\_BA}), since a full SVD on the
$d \times d$ reconstruction is prohibitively expensive at 7B scale.

\paragraph{M1 (SVD-structured mutation).}
Preserve learned structure while perturbing it within the singular-value
basis. Compute $\Delta W = U \Sigma V^\top$, perturb $\Sigma$
multiplicatively with log-normal noise $\Sigma \gets \Sigma \odot
\exp(\epsilon\, z),\ z \sim \mathcal{N}(0, I_r)$, and apply first-order
Cayley near-identity rotations $R = I + \epsilon K$ with
$K = (M - M^\top)/2,\ M \sim \mathcal{N}(0, I_{r\times r})$ to both
$U$ and $V$. Refactor with the balanced split
$B' = U' \sqrt{\Sigma'},\ A' = \sqrt{\Sigma'}\,V'^{\top}$. Default
$\epsilon = 0.1$. Note that $R$ is only approximately orthogonal: at
higher strength the Cayley first-order approximation drifts.

\paragraph{M2 (layer-selective Gaussian).}
Uniformly sample a fraction $f$ of LoRA module slots without replacement
and add per-tensor adaptive Gaussian noise to their $A$ and $B$ factors:
$A \mathrel{+}= \mathcal{N}(0, (\epsilon \cdot \operatorname{std}(A))^2)$
and similarly for $B$. The remaining $1-f$ of modules are copied
verbatim. Defaults $\epsilon = 0.1,\ f = 0.33$. Coherent within each
selected module; leaves the majority of the adapter untouched.

\paragraph{M3 (component masking).}
SVD the effective delta, sample $\lceil \rho \cdot r \rceil$ random
singular indices, zero the corresponding singular values, and refactor.
Default $\rho = 0.3$. The indices are \emph{uniformly random}, not
bottom-$k$, so the dropped directions are not necessarily the
least-important; effective rank drops by $\rho$.

\paragraph{M4 (full Gaussian).}
Add per-tensor adaptive Gaussian noise to every $A$ and $B$ in the
adapter: $A \mathrel{+}= \mathcal{N}(0, (\epsilon \cdot
\operatorname{std}(A))^2)$. Default $\epsilon = 0.15$, deliberately
larger than \textbf{M2}. Narrow tensors receive proportionally less
noise because the scale is tied to each tensor's running standard
deviation.

\paragraph{M5 (NEFTune-style).}
Dimension-aware uniform perturbation on the input factor only,
adapted from NEFTune's \citep{jain2024neftune} embedding-space noise.
For $A \in \mathbb{R}^{L \times d}$ draw
$\eta \sim \mathrm{Uniform}\!\left(-\alpha/\sqrt{Ld},\ \alpha/\sqrt{Ld}\right)$
elementwise and set $A' = A + \eta$; $B$ is left untouched. Default
$\alpha = 10$, which gives per-element noise $\approx 0.03$ on
Qwen2.5-Coder-7B rank-32 LoRA.

\paragraph{M6 (rank perturbation).}
Structured rank collapse plus fine-grained jitter on the survivors,
inspired by DyLoRA \citep{valipour2023dylora}. SVD the effective delta,
zero the bottom-$k$ singular values, and multiplicatively perturb the
surviving top $r-k$ values by $\mathcal{N}(1, \sigma)$. Defaults
$k=2,\ \sigma=0.05$. Preserves top singular \emph{directions}; this is a
real rank collapse unlike \textbf{M3}'s random masking.

\paragraph{\texttt{copy\_parent} (identity control).}
Deepcopy of the parent state dict. Used as the ceiling of the
mutation experiment: any real mutation should land at or below the
first-step reward of a copy\_parent child under the same retrain
schedule. Also the sensor that caught the FSDP~v1 \texttt{load\_lora\_only}
silent-no-op bug during Phase~0, when every copy\_parent child was
starting at the base-model reward regardless of parent snapshot, it was
immediate evidence that the adapter was never reaching the rollout
engine.

\paragraph{X1 (DARE).}
Drop-and-rescale recipe of \citet{yu2024dare}. For each of the four
factors $\{A^{(1)}, A^{(2)}, B^{(1)}, B^{(2)}\}$ independently, build a
Bernoulli keep mask with probability $1-p$, rescale the survivors by
$1/(1-p)$ to preserve expectation, and average the two parents. Default
$p = 0.7$. Because drop-and-rescale is applied independently to $A$ and
$B$, the effective-delta expectation contains cross-parent interaction
terms ($\mathbb{E}[B_{\text{child}} A_{\text{child}}^\top] \neq
\tfrac{1}{2}\sum_i B^{(i)} A^{(i)\top}$); this is a property of the
original DARE formulation.

\paragraph{X2 (layer-wise crossover).}
For each LoRA module slot, a coin flip selects $(A, B)$ entirely from
parent~1 or entirely from parent~2. This preserves intra-module
coherence (an $A,B$ pair is always consistent) but breaks inter-module
coherence. Reminiscent of TIES-style sign-aligned mergers
\citep{yadav2023ties} at module granularity.

\paragraph{X3 (SVD subspace crossover).}
Mix principal and secondary singular subspaces. SVD both parents, draw
$k \sim \mathrm{Uniform}\{1,\dots,r-1\}$, and build the child by
concatenating columns: $U_{\text{child}} = [U^{(1)}_{:,:k} \mid
U^{(2)}_{:,k:r}]$, with $\Sigma_{\text{child}}$ and $V_{\text{child}}$
similarly composed. Reconstruct $B', A'$ from the composed factors. The
concatenated $U_{\text{child}}$ and $V_{\text{child}}$ are \emph{not
themselves orthonormal} because they mix columns from two different
orthonormal bases; the result is a valid rank-$r$ LoRA but not the SVD
of any single matrix.

\paragraph{X4 (extrapolative).}
Linear combination past parent~2 along the parent difference vector,
analogous to task arithmetic \citep{ilharco2023taskarithmetic} with a
coefficient beyond one. Sample $\eta \sim \mathrm{Uniform}(\eta_{\min},
\eta_{\max})$ once per call and set $A_{\text{child}} = A^{(1)} + \eta\,
(A^{(2)} - A^{(1)})$, similarly for $B$. Default $(\eta_{\min},
\eta_{\max}) = (1.0, 1.5)$, so the child always lies beyond parent~2
along the parent-difference direction.

\paragraph{X5 (task-arithmetic linear merge).}
Tensor-wise convex combination:
$\text{child}[k] = \alpha\, P^{(1)}[k] + (1-\alpha)\, P^{(2)}[k]$ applied
to every tensor. Default $\alpha = 0.5$. Deterministic (no \texttt{rng}
draws); this is the canonical linear merge of \citet{ilharco2023taskarithmetic}.

\paragraph{X6 (TIES merge).}
Sign-aligned merge of \citet{yadav2023ties}. Per tensor: (1)~\emph{trim}
by zeroing the lowest $\tau$ fraction of elements per parent by
magnitude, (2)~\emph{elect} the consensus sign as the sign of the sum
over trimmed parents, (3)~\emph{disjoint merge} by averaging only the
parents whose sign agrees with the elected sign at that position;
positions where no parent agrees output zero. Default $\tau = 0.2$.
Deterministic. For two parents this degenerates to picking the
larger-magnitude survivor when the signs disagree.

\paragraph{X7 (DELLA).}
Magnitude-proportional drop with DARE-style rescale, inspired by
DELLA-Merging \citep{deep2024della}. Per tensor and per parent
independently: flatten and rank by $|\cdot|$ descending, assign
drop probability $p(r) = \varepsilon + (1-\varepsilon)\,r/(n-1)$ so the
largest element has drop probability $\varepsilon$ and the smallest
has drop probability $1$, apply independent Bernoulli keeps, rescale
survivors by $1/(1-p(r))$, and average the two processed parents.
Default $\varepsilon = 0.1$. Uses \texttt{rng}, so different seeds
produce different children.

\paragraph{X8 (SLERP).}
Spherical linear interpolation on the flattened-tensor vector. Per
tensor, flatten both parents to $a, b$ and compute
$\cos\theta = \langle a, b\rangle/(\|a\|\|b\|)$ clipped to $(-1, 1)$.
The merged vector is
$\text{merged} = \tfrac{\sin((1-t)\theta)}{\sin\theta}\,a +
\tfrac{\sin(t\theta)}{\sin\theta}\,b$. Falls back to linear
interpolation when either norm or $\sin\theta$ is near-zero. Default
$t = 0.5$. Deterministic. Useful when a linear midpoint would overshrink
the result because the parents are nearly anti-parallel.

\paragraph{X9 (Fisher-weighted merge).}
Data-free Fisher-weighted averaging in the spirit of \citet{matena2022fisher}:
$\text{child} = (F^{(1)}\, p^{(1)} + F^{(2)}\, p^{(2)}) / (F^{(1)} +
F^{(2)} + \varepsilon)$ per element, with $F^{(i)} = (p^{(i)})^2$ as a
data-free proxy for the diagonal Fisher. The original formulation
requires a gradient pass on a calibration set; our $p^2$ proxy preserves
the large-magnitude-wins behaviour without that cost. Deterministic.

\paragraph{\texttt{linear\_0\_5} (plain-mean control).}
Wrapper around \textbf{X5} with $\alpha = 0.5$ hard-coded under a
distinct registry name. Serves as the crossover counterpart to
\texttt{copy\_parent}: any informative crossover should outperform
a plain tensor-wise midpoint average under identical retrain
conditions.

\paragraph{Which operators ship live.}
Only the first subset \{M1, M2, M3, M4, X1, X2, X3, X4\} is active inside
the live population described in \S\ref{sec:method_evo}. The remaining
nine operators (M5, M6, \texttt{copy\_parent}, X5, X6, X7, X8, X9,
\texttt{linear\_0\_5}) are characterised in isolation via the
\texttt{experiments/lora\_ops/} benchmark (see \S\ref{app:loraops})
but did not enter the main self-play loop.

\clearpage
\section{Sample Generated Problems over Training}
\label{app:samples}

\begin{figure}[ht!]
\centering
\input{figs_text/problem_evolution.tex}
\caption{\textbf{Generator outputs at matched training steps.} Left:
baseline AZR; right: one PopuLoRA teacher. By step~100 the baseline
has collapsed to \texttt{return number * 3}, while the population is
still producing programs with branching and state.}
\label{fig:problem_evolution}
\end{figure}
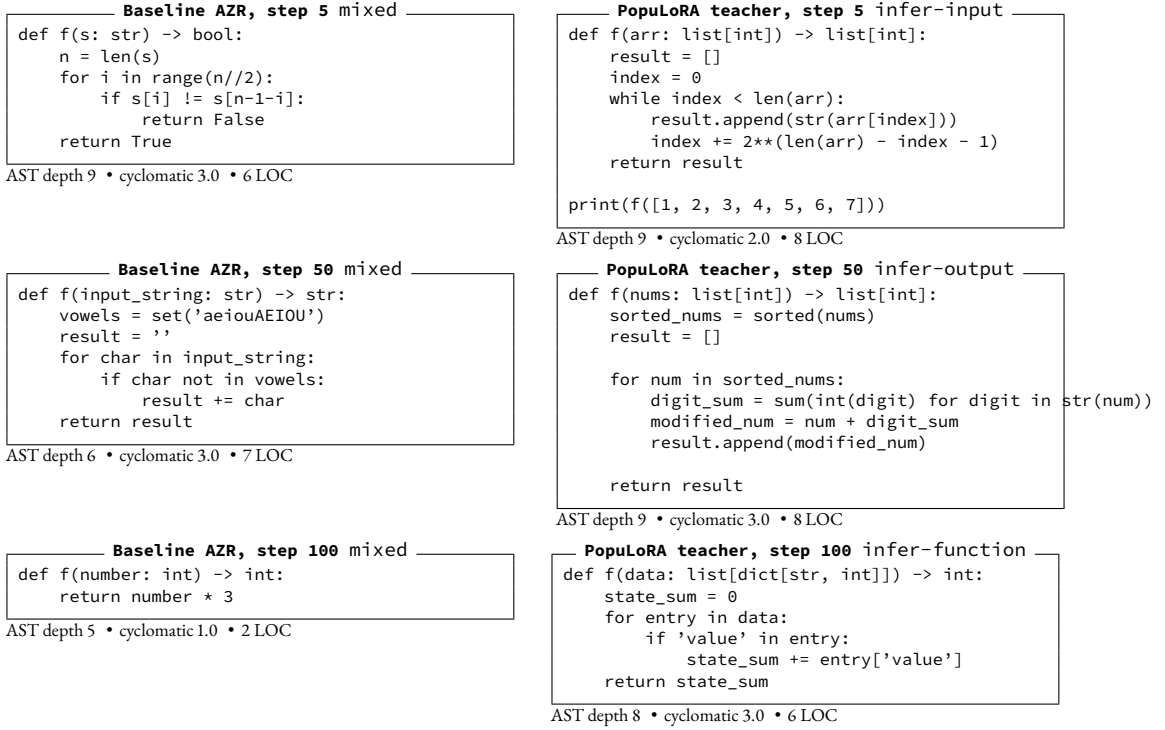

Figure~\ref{fig:problem_evolution} pairs one baseline-generated and one
population-generated problem from matched training steps, drawn from
the saved per-step problem archives. Picks are deterministic: at each
step we take a median-complexity quality-1.0 problem, subject to a
loose line-count bound so snippets fit the figure; at step 100 we
additionally report the most trivial quality-1.0 baseline problem to
illustrate the mode-collapse endpoint the baseline drifts into. Across
all three rows, the baseline's outputs shrink to increasingly vacuous
programs, while the population continues producing substantive code
across the three AZR task types.

\section{Problem-Space Coverage}
\label{app:coverage}

\paragraph{Structural complexity metrics.} The four metrics tracked
in \S\ref{sec:exp_modecollapse} (and used as the descriptor for the
CVT archive below) are computed from the Python AST of each
teacher-generated program. \textbf{AST depth} is the maximum nesting
depth of the parsed syntax tree. \textbf{Cyclomatic complexity} is
McCabe's count of linearly independent paths through the program,
computed as $1 + \#\{\texttt{if}, \texttt{elif}, \texttt{for},
\texttt{while}, \texttt{except}, \texttt{and}, \texttt{or},
\texttt{assert}\}$. \textbf{Lines of code} counts non-blank,
non-comment source lines, and \textbf{variable count} is the number
of distinct identifiers introduced as variables.

\begin{figure}[ht!]
\centering
\includegraphics[width=0.6\linewidth]{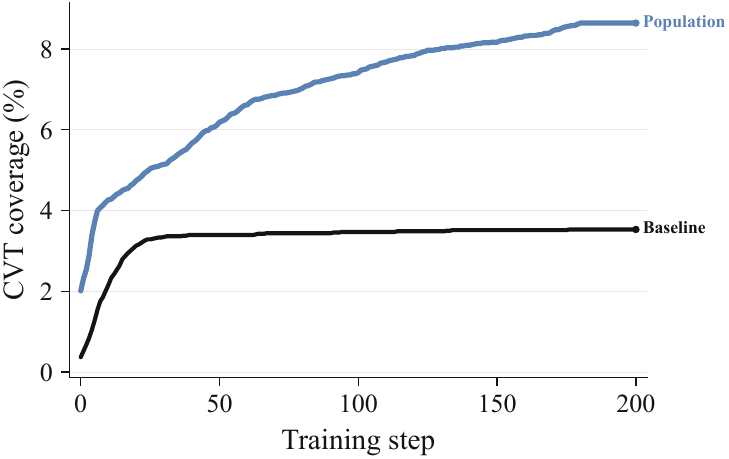}
\caption{\textbf{Problem-space coverage.} CVT archive grid coverage
(percent of the 4096-cell budget). Baseline (black) vs population
(blue).}
\label{fig:coverage}
\end{figure}

Figure~\ref{fig:coverage} plots the fraction of CVT archive grid tiles
filled by validated teacher-generated problems, as training progresses.
We report this quantity as a percent: population coverage is read
directly from \texttt{archive/coverage\_pct}, and the baseline proxy
divides its running \texttt{archive/total\_problems} by the
grid-cell budget of 4096 tiles. Baseline coverage plateaus early and
stays flat; population coverage keeps expanding through the full
200 steps, consistent with the complexity-growth picture in the main
text (\S\ref{sec:exp_modecollapse}).

\section{Per-Type Training Dynamics}
\label{app:pertype}

\begin{figure}[ht!]
\centering
\includegraphics[width=\linewidth]{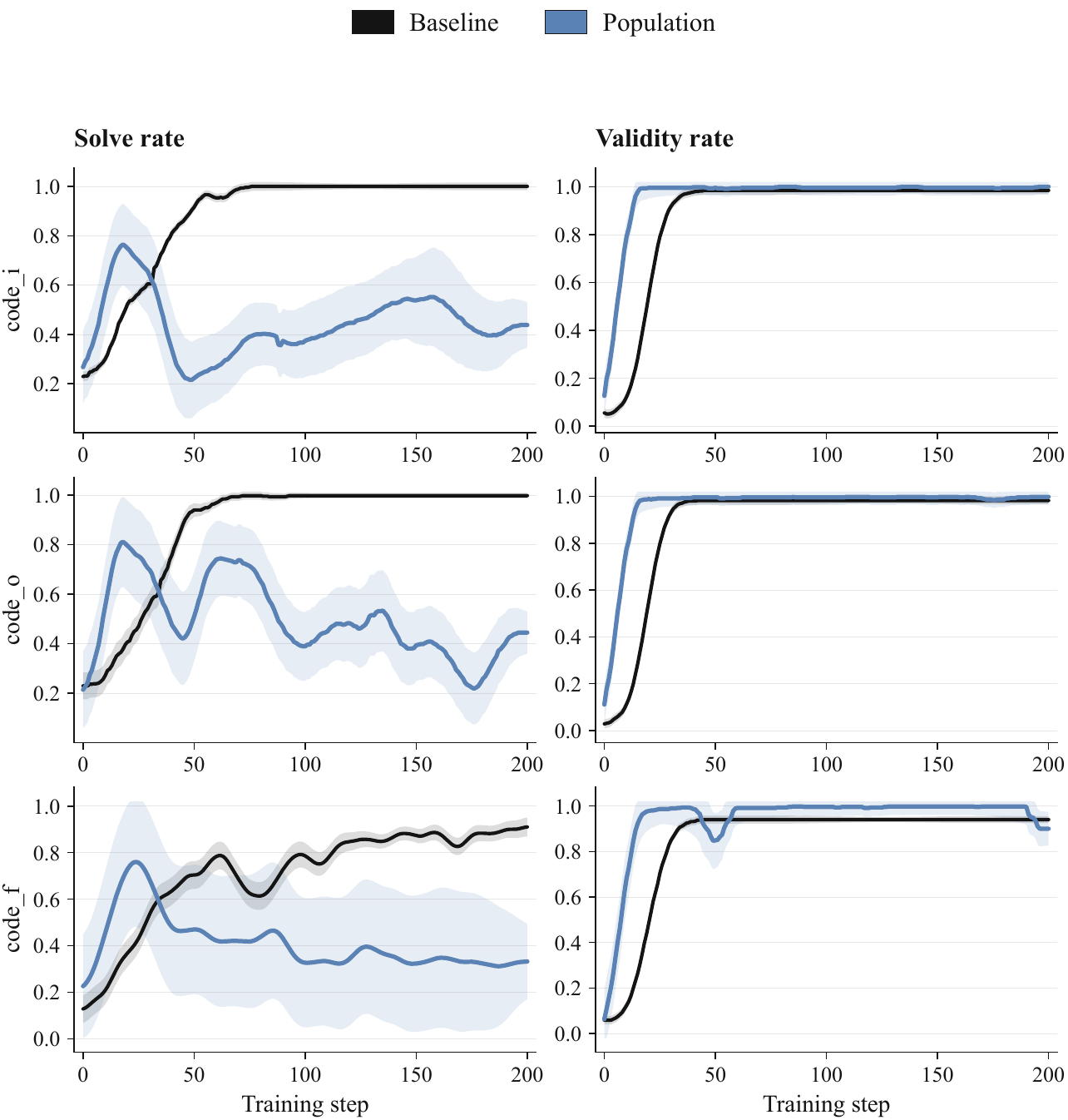}
\caption{\textbf{Per-type breakdown of Figure~\ref{fig:dynamics}.}
Rows: \texttt{code\_i}, \texttt{code\_o}, \texttt{code\_f}.
Columns: solve rate, validity rate. Baseline (black) vs.\ population
mean (blue) with per-member spread.}
\label{fig:dynamics_per_type}
\end{figure}

Figure~\ref{fig:dynamics_per_type} disaggregates the main-text
training dynamics (Figure~\ref{fig:dynamics}) by AZR task type.

\clearpage
\section{Solve Rate by Problem Type}
\label{app:solverate_by_type}

\begin{figure}[ht!]
\centering
\includegraphics[width=\linewidth]{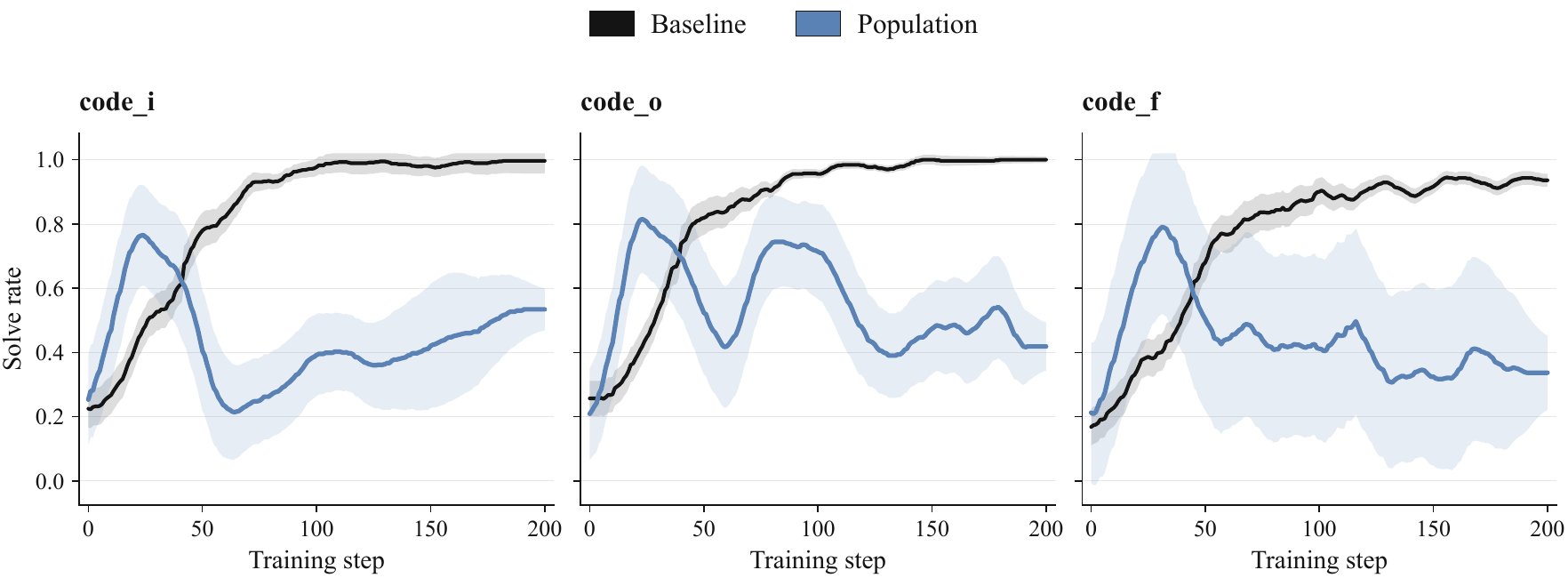}
\caption{\textbf{Solve rate by problem type.} Baseline (black) vs.\
population (blue) with per-member spread. The baseline saturates on
all three types; the population oscillates as teachers co-adapt.}
\label{fig:solve_rate_by_type}
\end{figure}

Figure~\ref{fig:solve_rate_by_type} isolates the solver's solve rate
for each of the three AZR task types. The baseline reaches near-perfect
solve rate on all three types, consistent with self-calibration to easy
problems. The population's solve rate oscillates on each type, with
the oscillation frequency varying across types (fastest on output
prediction, slowest on induction), matching the per-type dynamics in
Figure~\ref{fig:dynamics_per_type}.

\section{Cross-Evaluation over Training}
\label{app:crosseval}

\begin{figure}[ht!]
\centering
\includegraphics[width=\linewidth]{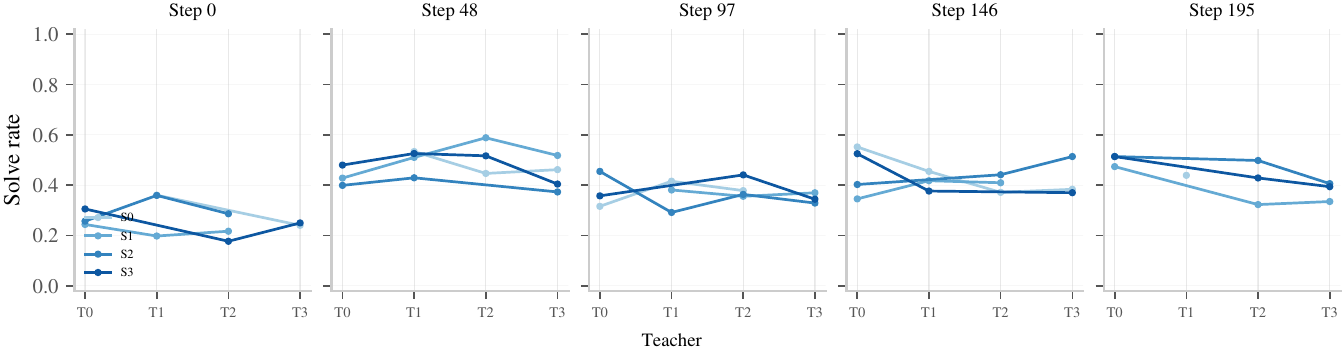}
\caption{\textbf{Cross-evaluation over training.} Parallel
coordinates at five equispaced snapshots. Each vertical axis is
one teacher; each line is one student's solve rate against that
teacher. Flat lines indicate uniform solvers; kinks indicate
specialisation against particular teachers.}
\label{fig:crosseval}
\end{figure}

Figure~\ref{fig:crosseval} shows per-student solve-rate profiles
against each teacher at five equispaced training snapshots.

\clearpage
\section{Extended Downstream Evaluation}
\label{sec:app_full_ft_comparison}

In the main paper (Figure~\ref{fig:downstream}) we report both the
population mean and the per-benchmark best for each role. Table~\ref{tab:downstream}
provides the numeric complement to the figure.

\begin{table}[h]
\centering
\resizebox{\textwidth}{!}{%
\begin{tabular}{l *{3}{r} @{\hskip 4pt} *{7}{r} @{\hskip 4pt} *{3}{r}}
\toprule
 & \multicolumn{3}{c}{Code} & \multicolumn{7}{c}{Math} & \multicolumn{3}{c}{Aggregate} \\
\cmidrule(lr){2-4} \cmidrule(lr){5-11} \cmidrule(l){12-14}
Method & HE+ & MBPP+ & LCB & AIME24 & AIME25 & AMC & M-500 & Min. & GSM & Oly. & Code & Math & All \\
\midrule
Qwen2.5-Coder-7B & 79.3 & 68.3 & 14.3 & 6.7 & 3.3 & 37.5 & 55.0 & 15.8 & 83.8 & 23.7 & 53.9 & 32.3 & 38.8 \\
Baseline AZR (LoRA) & 79.0{\scriptsize$\pm$2.1} & 68.1{\scriptsize$\pm$1.2} & 16.5{\scriptsize$\pm$0.8} & 8.3{\scriptsize$\pm$1.7} & 3.4{\scriptsize$\pm$3.4} & 42.5{\scriptsize$\pm$5.0} & 65.5{\scriptsize$\pm$6.1} & 23.3{\scriptsize$\pm$4.2} & 85.6{\scriptsize$\pm$3.7} & 31.3{\scriptsize$\pm$4.0} & 54.5{\scriptsize$\pm$1.4} & 37.1{\scriptsize$\pm$4.0} & 42.3{\scriptsize$\pm$3.2} \\
\midrule
1T+1S Teacher & 81.7 & 70.9 & 16.6 & 0.0 & 3.3 & \textbf{55.0} & 72.2 & 23.9 & 89.0 & 36.0 & 56.4 & 39.9 & 44.9 \\
1T+1S Student & \textbf{82.9} & \textbf{71.2} & 17.3 & 6.7 & 3.3 & 50.0 & \textbf{73.8} & 22.8 & \textbf{89.5} & 36.4 & 57.1 & 40.4 & 45.4 \\
\midrule
4T+4S Teacher & 80.2{\scriptsize$\pm$0.8} & 69.0{\scriptsize$\pm$0.4} & 27.1{\scriptsize$\pm$0.3} & 5.8{\scriptsize$\pm$1.7} & 4.2{\scriptsize$\pm$1.7} & 43.1{\scriptsize$\pm$2.4} & 69.8{\scriptsize$\pm$0.3} & 29.8{\scriptsize$\pm$1.5} & 84.2{\scriptsize$\pm$0.2} & 35.8{\scriptsize$\pm$1.3} & 58.8{\scriptsize$\pm$0.5} & 39.0{\scriptsize$\pm$1.3} & 44.9{\scriptsize$\pm$1.0} \\
4T+4S Student & 80.2{\scriptsize$\pm$0.4} & 69.2{\scriptsize$\pm$0.6} & 27.4{\scriptsize$\pm$0.1} & 5.8{\scriptsize$\pm$3.2} & 6.7{\scriptsize$\pm$2.7} & 45.6{\scriptsize$\pm$2.4} & 70.0{\scriptsize$\pm$0.7} & 30.5{\scriptsize$\pm$1.0} & 84.6{\scriptsize$\pm$0.5} & 35.1{\scriptsize$\pm$1.2} & 58.9{\scriptsize$\pm$0.3} & 39.8{\scriptsize$\pm$1.7} & 45.5{\scriptsize$\pm$1.3} \\
4T+4S Worst & 79.3 & 68.5 & 26.9 & 3.3 & 3.3 & 40.0 & 69.0 & 28.7 & 83.8 & 33.6 & 58.2 & 37.4 & 43.6 \\
4T+4S Best & 81.1 & 69.8 & \textbf{27.5} & \textbf{10.0} & \textbf{10.0} & 47.5 & 70.8 & \textbf{32.0} & 85.0 & \textbf{37.3} & \textbf{59.5} & \textbf{41.8} & \textbf{47.1} \\
\midrule
8T+8S Teacher & 80.6{\scriptsize$\pm$0.6} & 68.8{\scriptsize$\pm$0.5} & 21.1{\scriptsize$\pm$0.3} & 4.6{\scriptsize$\pm$1.7} & 4.2{\scriptsize$\pm$1.5} & 38.1{\scriptsize$\pm$4.8} & 67.0{\scriptsize$\pm$0.7} & 22.2{\scriptsize$\pm$1.1} & 87.0{\scriptsize$\pm$0.4} & 32.6{\scriptsize$\pm$1.0} & 56.8{\scriptsize$\pm$0.5} & 36.5{\scriptsize$\pm$1.6} & 42.6{\scriptsize$\pm$1.3} \\
8T+8S Student & 80.9{\scriptsize$\pm$0.3} & 68.7{\scriptsize$\pm$0.5} & 20.9{\scriptsize$\pm$0.4} & 7.1{\scriptsize$\pm$1.2} & 5.0{\scriptsize$\pm$2.5} & 34.7{\scriptsize$\pm$3.4} & 66.6{\scriptsize$\pm$0.9} & 21.8{\scriptsize$\pm$1.4} & 86.8{\scriptsize$\pm$0.4} & 32.9{\scriptsize$\pm$0.5} & 56.8{\scriptsize$\pm$0.4} & 36.4{\scriptsize$\pm$1.5} & 42.5{\scriptsize$\pm$1.2} \\
8T+8S Worst & 79.9 & 68.0 & 20.3 & 3.3 & 0.0 & 30.0 & 64.8 & 20.2 & 85.9 & 31.1 & 56.1 & 33.6 & 40.4 \\
8T+8S Best & 81.1 & 69.6 & 21.7 & \textbf{10.0} & 6.7 & 45.0 & 68.2 & 24.6 & 87.6 & 33.9 & 57.5 & 39.4 & 44.8 \\
\bottomrule
\end{tabular}}
\caption{\textbf{Downstream pass@1 (\%).} Numeric complement to
Figure~\ref{fig:downstream}. Mean $\pm$ std across adapters within
each role. Worst/Best are the single weakest/strongest adapter
(per-benchmark selection on the test set).
Best value per column in \textbf{bold}.
The 4T+4S Worst adapter (43.6\% aggregate) still outperforms the
baseline (42.3\%), indicating that co-evolution lifts the entire
population rather than concentrating gains in a few members.}
\label{tab:downstream}
\end{table}

Here we provide two complementary views.

\begin{figure}[ht!]
\centering
\includegraphics[width=\linewidth]{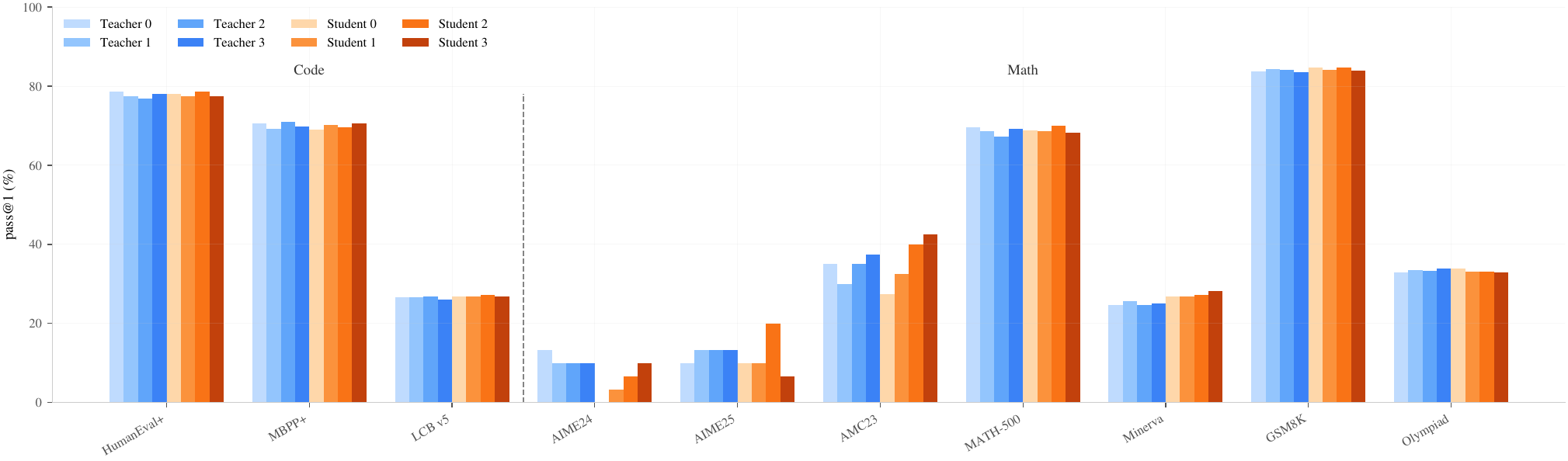}
\caption{Pass@1 for each of the 4 teachers and 4 students from the
4T+4S population. The main text (Figure~\ref{fig:downstream})
reports the population mean and per-benchmark best; here we show all
individual adapters.}
\label{fig:a_pop4t4s_all_adapters_bars}
\end{figure}

\begin{figure}[ht!]
\centering
\includegraphics[width=\linewidth]{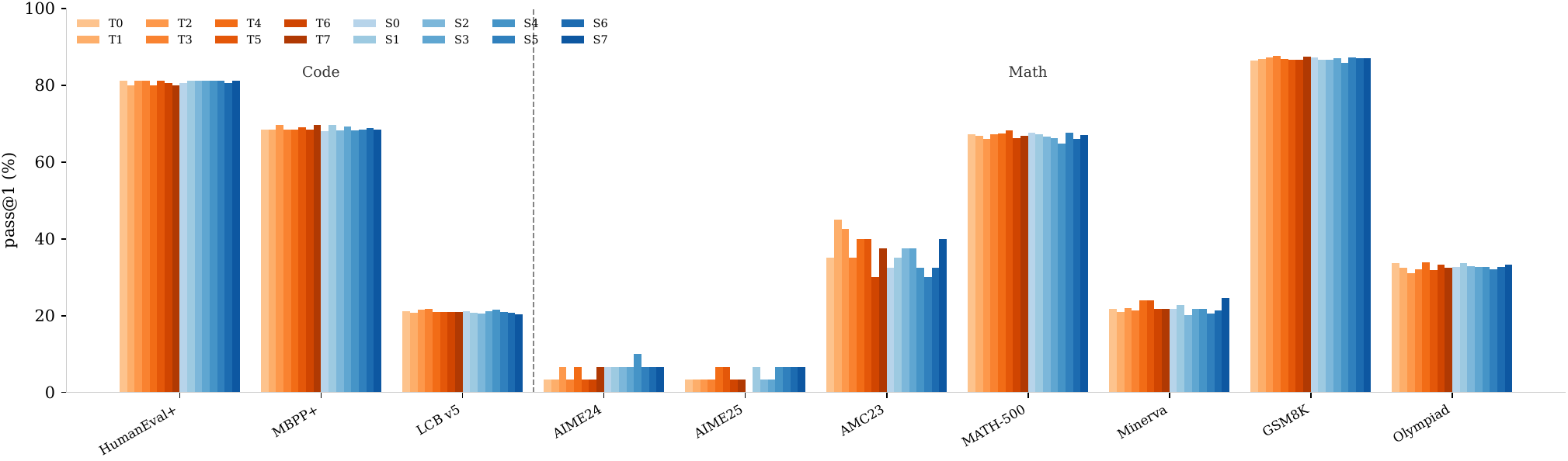}
\caption{Pass@1 for each of the 8 teachers and 8 students from the
8T+8S population.}
\label{fig:a_pop8t8s_all_adapters_bars}
\end{figure}

\paragraph{Per-adapter breakdown.}
Figures~\ref{fig:a_pop4t4s_all_adapters_bars}
and~\ref{fig:a_pop8t8s_all_adapters_bars} disaggregate the 4T+4S and
8T+8S populations into individual adapters. The per-adapter view shows
that all population members are fairly competent across the board: the
spread within each role is modest, indicating that co-evolution
produces a population of generally strong reasoners rather than narrow
specialists.

\begin{figure}[ht!]
\centering
\includegraphics[width=\linewidth]{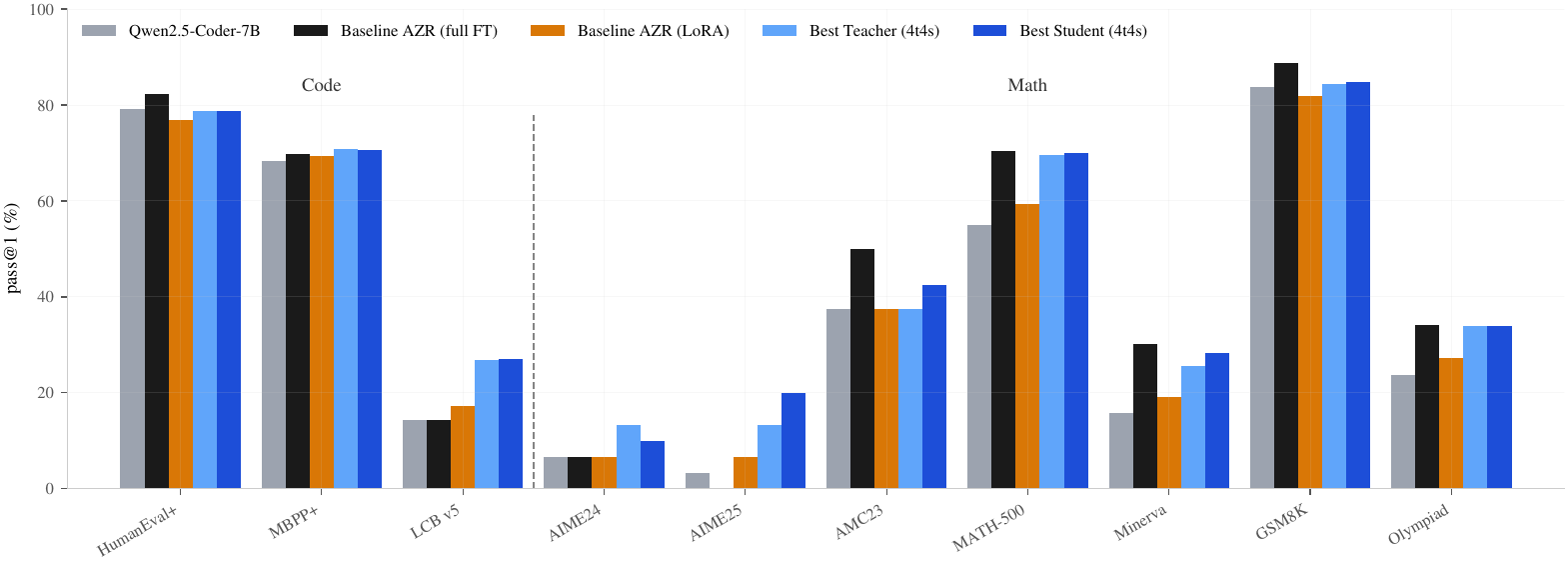}
\caption{Downstream pass@1 including the full-finetune Baseline AZR
(300 gradient steps, non-LoRA). Compare with
Figure~\ref{fig:downstream}, which uses only the per-adapter compute-matched LoRA
baseline.}
\label{fig:downstream_full_ft}
\end{figure}

\paragraph{Comparison with full-finetune AZR.}
Figure~\ref{fig:downstream_full_ft} adds the publicly released
Baseline AZR checkpoint \citep{zhao2025absolute}, which was produced by
full fine-tuning (not LoRA) and trained for 300 gradient steps,
50\% more than the 200 steps used by both the Baseline AZR (LoRA) and
the PopuLoRA population members. The comparison is therefore not
per-adapter compute-matched: the full-finetune baseline has both more parameter
updates and full-rank gradient access. Despite this advantage, the
best PopuLoRA student matches or exceeds the full-finetune baseline on
the majority of benchmarks, particularly on LiveCodeBench and the
competition-level math benchmarks (AIME, AMC).

\clearpage
\section{Full \texttt{lora\_ops} Benchmark Results}
\label{app:loraops}

The figures below come from the isolated operator-evaluation sweep at
\texttt{experiments/lora\_ops/}. For each operator, we apply it to a frozen
parent adapter at five snapshot steps (5, 10, 25, 50, 100); the child
then re-trains for 50 steps on the parent's task and its pass-rate curve
is recorded. The main-paper Figure~\ref{fig:retention} shows the subset
of four mutation and four crossover operators that we ship in the live
population; here we report the full operator catalog.

\begin{figure}[ht!]
\centering
\includegraphics[width=\linewidth]{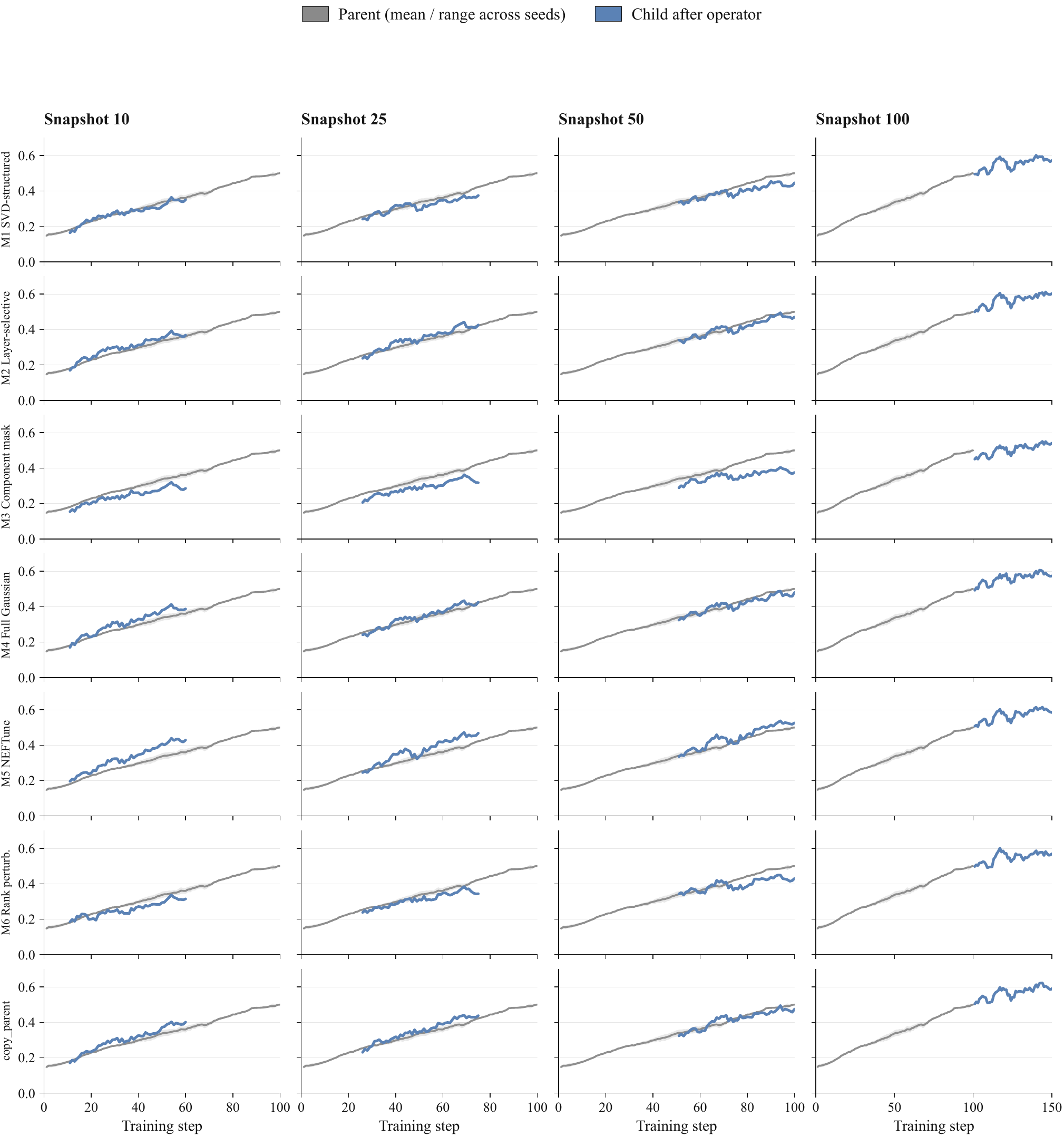}
\caption{\textbf{Mutation-operator retention across snapshot steps.}
Rows: mutation operators M1--M6 plus \texttt{copy\_parent} control.
Columns: snapshot steps (10, 25, 50, 100). Parent's 100-step
learning curve is drawn
in grey, and the child's 50-step retraining curve in colour, with the
child's x-axis offset by the snapshot step so both live on the same
global-step scale.}
\end{figure}

\begin{figure}[ht!]
\centering
\includegraphics[width=\linewidth]{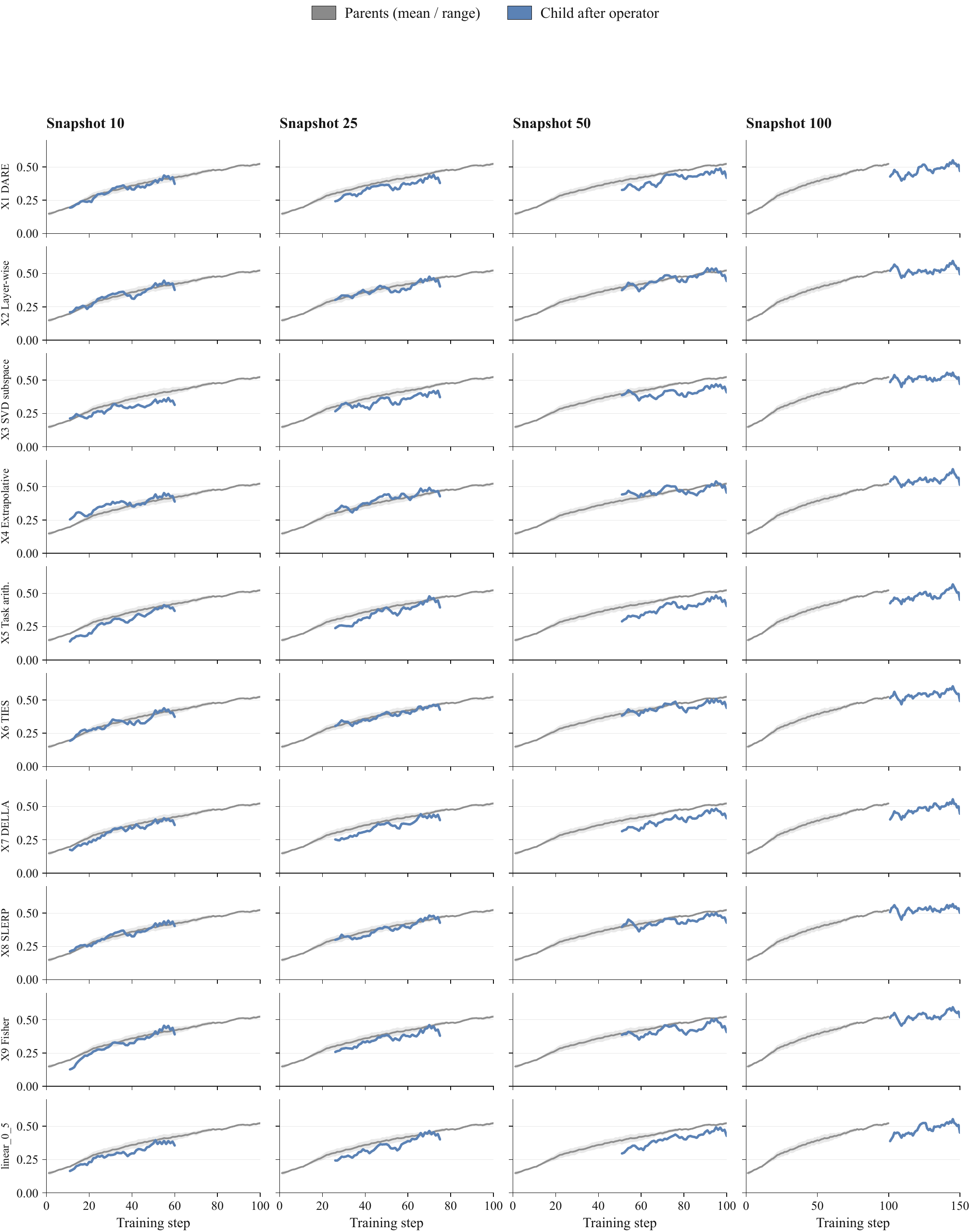}
\caption{\textbf{Crossover-operator retention across snapshot steps.}
Same layout as the mutation figure: rows are X1--X9 plus the
\texttt{linear\_0\_5} plain-average control; columns are snapshot
steps (10, 25, 50, 100). Parents from \texttt{exp\_c1} task-merging
sweep in grey, child retraining in colour
with the snapshot-step offset.}
\end{figure}

\clearpage
\section{Training Diagnostics}
\label{app:diagnostics}

\begin{figure}[ht!]
\centering
\includegraphics[width=\linewidth]{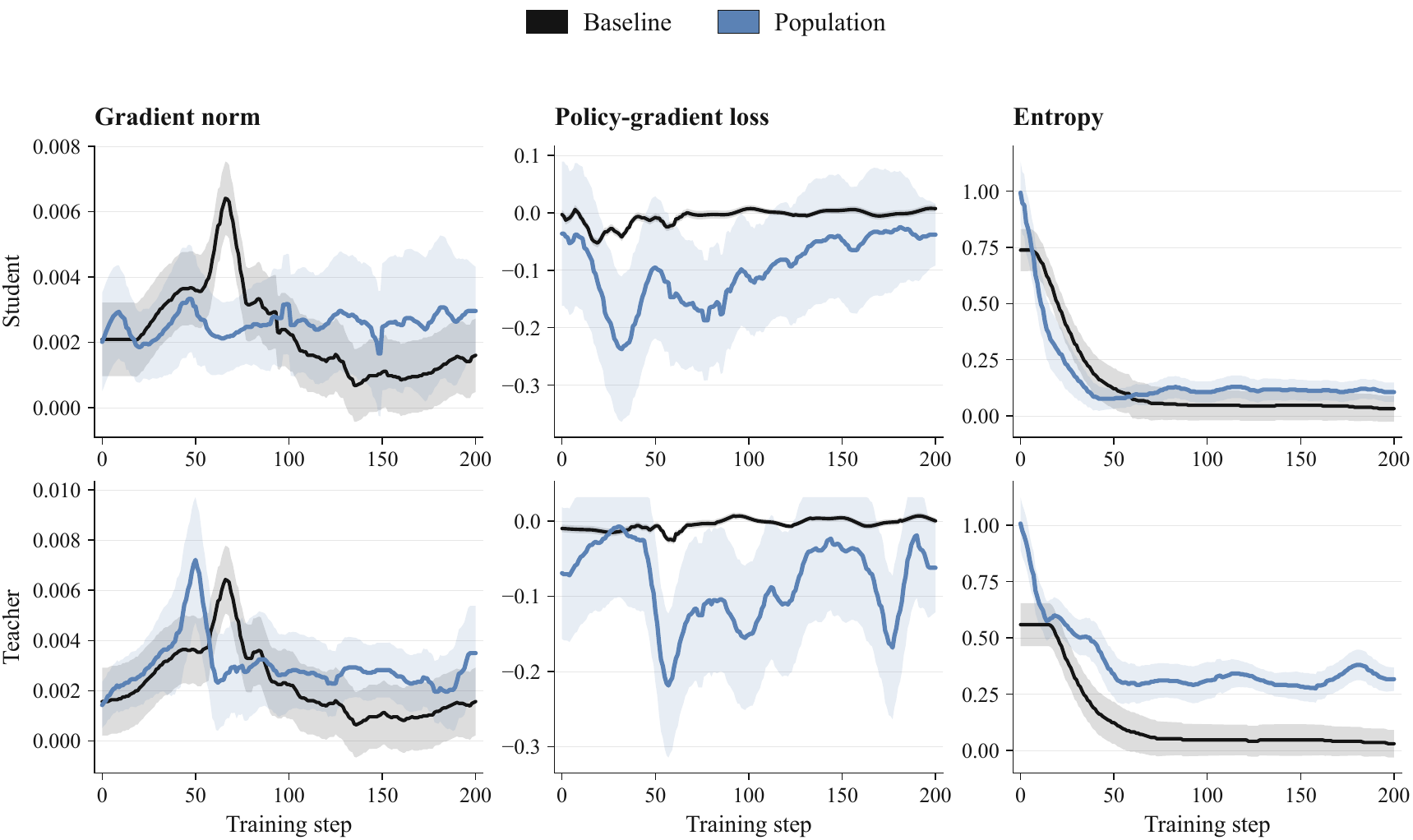}
\caption{\textbf{Training diagnostics.} Gradient norm,
policy-gradient loss, entropy. Baseline (black) vs population mean
(blue) with per-member spread (light blue).}
\label{fig:diagnostics}
\end{figure}

Figure~\ref{fig:diagnostics} shows three standard policy-gradient
diagnostics for both arms: actor gradient norm, policy-gradient loss,
and entropy of the actor's output distribution. All three metrics
track within the usual range for stable training; the population's
per-member spread reflects the independent adapters training in
parallel. We omit a KL panel because we train without a KL penalty
($\texttt{kl\_loss\_coef}=0$, $\texttt{kl\_ctrl.kl\_coef}=0$), so the
KL term is identically zero for every step of both arms.

\section{Response Length over Training}
\label{app:response_length}

\begin{figure}[ht!]
\centering
\includegraphics[width=0.55\linewidth]{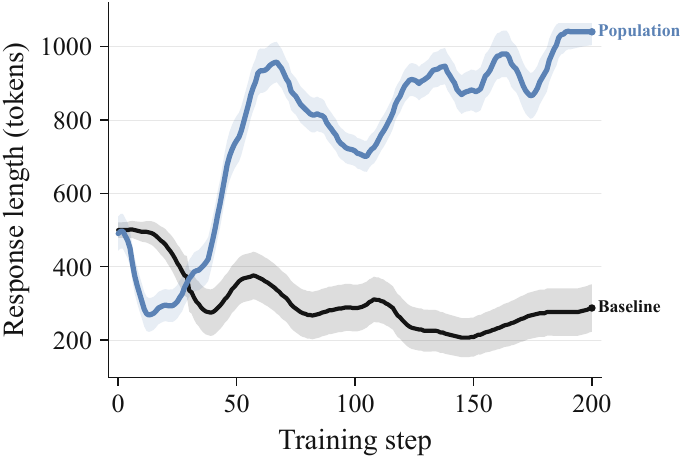}
\caption{\textbf{Response length over training.} Baseline (black)
collapses to short responses (${\sim}250$ tokens); population (blue)
grows to ${\sim}1000$ tokens as problem complexity increases.}
\label{fig:response_length}
\end{figure}

Figure~\ref{fig:response_length} tracks the mean response length (in
tokens) over training. The baseline's responses shorten steadily as the
agent self-calibrates to trivial problems that require only short
programs and short solutions. The population shows the opposite trend:
response length grows throughout training, reaching roughly $3{\times}$
the baseline's by the end of the run. Longer responses have been
associated with more elaborate reasoning in prior RLVR
work~\citep{guo2025deepseek}; here the effect is driven by teachers
generating increasingly complex problems that demand longer programs
and more detailed solutions.

%% file: figs_text/problem_evolution.tex
% Auto-generated by scripts/cherry_pick_problems.py
% Paired baseline vs population generator outputs at matched training steps.

% --- step 5 ---
\begin{minipage}[t]{0.48\linewidth}
\begin{Verbatim}[fontsize=\scriptsize, frame=single, framesep=4pt, label={\textbf{Baseline AZR, step 5} \hfill {\footnotesize mixed}}, labelposition=topline, samepage=true]
def f(s: str) -> bool:
    n = len(s)
    for i in range(n//2):
        if s[i] != s[n-1-i]:
            return False
    return True
\end{Verbatim}
\vspace{-6pt}
{\scriptsize AST depth 9 \, \textbullet\, cyclomatic 3.0 \, \textbullet\, 6 LOC}
\end{minipage}%
\hfill
\begin{minipage}[t]{0.48\linewidth}
\begin{Verbatim}[fontsize=\scriptsize, frame=single, framesep=4pt, label={\textbf{PopuLoRA teacher, step 5} \hfill {\footnotesize infer-input}}, labelposition=topline, samepage=true]
def f(arr: list[int]) -> list[int]:
    result = []
    index = 0
    while index < len(arr):
        result.append(str(arr[index]))
        index += 2**(len(arr) - index - 1)
    return result

print(f([1, 2, 3, 4, 5, 6, 7]))
\end{Verbatim}
\vspace{-6pt}
{\scriptsize AST depth 9 \, \textbullet\, cyclomatic 2.0 \, \textbullet\, 8 LOC}
\end{minipage}

\vspace{0.8em}

% --- step 50 ---
\begin{minipage}[t]{0.48\linewidth}
\begin{Verbatim}[fontsize=\scriptsize, frame=single, framesep=4pt, label={\textbf{Baseline AZR, step 50} \hfill {\footnotesize mixed}}, labelposition=topline, samepage=true]
def f(input_string: str) -> str:
    vowels = set('aeiouAEIOU')
    result = ''
    for char in input_string:
        if char not in vowels:
            result += char
    return result
\end{Verbatim}
\vspace{-6pt}
{\scriptsize AST depth 6 \, \textbullet\, cyclomatic 3.0 \, \textbullet\, 7 LOC}
\end{minipage}%
\hfill
\begin{minipage}[t]{0.48\linewidth}
\begin{Verbatim}[fontsize=\scriptsize, frame=single, framesep=4pt, label={\textbf{PopuLoRA teacher, step 50} \hfill {\footnotesize infer-output}}, labelposition=topline, samepage=true]
def f(nums: list[int]) -> list[int]:
    sorted_nums = sorted(nums)
    result = []

    for num in sorted_nums:
        digit_sum = sum(int(digit) for digit in str(num))
        modified_num = num + digit_sum
        result.append(modified_num)

    return result
\end{Verbatim}
\vspace{-6pt}
{\scriptsize AST depth 9 \, \textbullet\, cyclomatic 3.0 \, \textbullet\, 8 LOC}
\end{minipage}

\vspace{0.8em}

% --- step 100 ---
\begin{minipage}[t]{0.48\linewidth}
\begin{Verbatim}[fontsize=\scriptsize, frame=single, framesep=4pt, label={\textbf{Baseline AZR, step 100} \hfill {\footnotesize mixed}}, labelposition=topline, samepage=true]
def f(number: int) -> int:
    return number * 3
\end{Verbatim}
\vspace{-6pt}
{\scriptsize AST depth 5 \, \textbullet\, cyclomatic 1.0 \, \textbullet\, 2 LOC}
\end{minipage}%
\hfill
\begin{minipage}[t]{0.48\linewidth}
\begin{Verbatim}[fontsize=\scriptsize, frame=single, framesep=4pt, label={\textbf{PopuLoRA teacher, step 100} \hfill {\footnotesize infer-function}}, labelposition=topline, samepage=true]
def f(data: list[dict[str, int]]) -> int:
    state_sum = 0
    for entry in data:
        if 'value' in entry:
            state_sum += entry['value']
    return state_sum
\end{Verbatim}
\vspace{-6pt}
{\scriptsize AST depth 8 \, \textbullet\, cyclomatic 3.0 \, \textbullet\, 6 LOC}
\end{minipage}